\documentclass[final,5p,times,twocolumn,authoryear]{elsarticle}

\usepackage{lineno}
\usepackage[hidelinks]{hyperref}
\usepackage{url}
\usepackage{siunitx}
\usepackage{enumitem}
\usepackage{booktabs}
\usepackage{graphicx}
\usepackage{lscape}
\usepackage{multirow}
\usepackage{longtable,tabu}
\usepackage{perpage}
\usepackage{makecell}
\MakePerPage{footnote}
\usepackage{float}

\begin{document}


\begin{frontmatter}



\title{Convolutional neural networks for mineral prospecting through alteration mapping with remote sensing data}


\author[fn1]{Ehsan Farahbakhsh\corref{cor1}}
\fntext[fn1]{EarthByte Group, School of Geosciences, The University of Sydney, Sydney, Australia}
\ead{e.farahbakhsh@sydney.edu.au}
\cortext[cor1]{Corresponding author}

\author[fn2]{Dakshi Goel}
\fntext[fn2]{Department of Electrical Engineering, Indian Institute of Technology Jammu, India}

\author[fn3]{Dhiraj Pimparkar}
\fntext[fn3]{Department of Mechanical Engineering, Indian Institute of Technology Jammu, India}

\author[fn1]{R. Dietmar M\"uller}

\author[fn4]{Rohitash Chandra}
\fntext[fn4]{Transitional Artificial Intelligence Research Group, School of Mathematics and Statistics, University of New South Wales, Sydney, Australia}


\begin{abstract}

Traditional geological mapping methods, which rely on field observations and rock sample analysis, are inefficient for continuous spatial mapping of geological features such as alteration zones. Deep learning models such as convolutional neural networks (CNNs) have ushered in a transformative era in remote sensing data analysis. CNNs excel in automatically extracting features from image data for classification and regression problems. CNNs have the ability to pinpoint specific mineralogical changes attributed to mineralisation processes by discerning subtle features within remote sensing data. In this study, we deploy CNNs with three sets of remote sensing data, namely Landsat 8, Landsat 9, and ASTER, to delineate diverse alteration zones within a mineral-rich region north of Broken Hill in western New South Wales, Australia. Our methodology involves model training using two distinct sets of training samples generated through ground truth data and a fully automated approach through selective principal component analysis (PCA). We also compare CNNs with conventional machine learning models, including k-nearest neighbours, support vector machines, and multilayer perceptron. Our findings indicate that training with a ground truth-based dataset produces more reliable alteration maps. Additionally, we find that CNNs perform slightly better when compared to conventional machine learning models, which further demonstrates the ability of CNNs to capture spatial patterns in remote sensing data effectively. We find that Landsat 9 surpasses Landsat 8 in mapping iron oxide areas when employing the CNNs model trained with ground truth data obtained by field surveys. We also observe that using ASTER data with the CNNs model trained on the ground truth-based dataset produces the most accurate maps for two other important types of alteration zones, argillic and propylitic. This underscores the utility of CNNs in enhancing the efficiency and precision of geological mapping, particularly in discerning subtle alterations indicative of mineralisation processes, especially those associated with critical metal resources.

\end{abstract}


\end{frontmatter}


\section{Introduction}

Geological maps are traditionally crafted through ground surveys and founded on field observations. They frequently incur inevitable errors due to the lack of spatial continuity of the field observations, thus yielding inaccurate representations \citep{campbell2005analysis}. Recognising these limitations, geologists have been prompted to seek innovative approaches and efficient methodologies to accurately map geological features, particularly alteration zones \citep{kesler2007mineral,mccuaig2010translating}. The utilisation of remote sensing data for alteration mapping emerges as a pivotal technique in regional mineral exploration, enabling the precise spatial identification of alteration zones associated with mineralisation processes \citep{mohamed2021geological}. Over the years, alteration mapping via remote sensing has garnered considerable attention due to its capacity to cover expansive Earth surfaces efficiently, facilitating the pinpointing of areas most likely to harbour economically viable mineral resources \citep{sabins1999remote,rowan2003lithologic,mohamed2021geological}. Remote sensing data sourced from airborne and satellite platforms constitute a key asset in this endeavour \citep{shirmard2022review}. Among different types of sensors for remote sensing, optical sensors stand out as the most commonly employed and are proficient at detecting changes in the reflectance and absorption characteristics of minerals on the Earth's surface \citep{van2012multi}. These changes signify various forms of alteration, including hydrothermal alteration induced by the circulation of hot fluids through rocks, and supergene alteration resulting from surface weathering processes \citep{chirico2022mapping}. Adopting alteration mapping with remote sensing data not only diminishes the time and cost associated with mineral exploration but also reduces the environmental footprint by narrowing down the areas requiring exploration. Furthermore, alteration mapping enhances the precision of mineral exploration by furnishing detailed insights into the location and extent of mineralisation processes \cite{maleki2021hydrothermal}.

The conventional approaches employed in processing remote sensing data for alteration mapping typically entail using spectral indices or basic classification techniques. Although these methods have widespread applications, they also face certain limitations. Spectral indices hinge on predefined equations based on the spectral properties of specific minerals, assuming that their signatures remain constant across diverse areas, unaffected by external factors such as topography, soil, or vegetation \citep{vicente2011identification}. However, the spectral signatures of minerals exhibit considerable variability due to diverse factors, rendering spectral indices prone to inaccuracies if these variations are not duly considered \citep{van2012multi}.

The effectiveness of classification techniques, such as logistic regression, which are founded on simple principles, largely depends on the user's choice of training data samples \citep{lyons2018comparison}. Inaccuracies in the resulting classification map may arise if the chosen training samples fail to represent the actual classes accurately. Moreover, these methods exclusively leverage spectral information for pixel classification, neglecting valuable spatial information that could aid in distinguishing between different classes \citep{shirmard2022review}. Additionally, conventional classification methods are susceptible to noise and outliers in the data, potentially compromising the accuracy of the resultant classification map \citep{lyons2018comparison}. Furthermore, these methods presuppose linear and additive relationships between different spectral bands, an assumption that may not hold true for the intricate mineral assemblages commonly encountered in alteration zones \citep{shirmard2022review}. These limitations underscore the imperative for more advanced techniques, such as machine learning algorithms, to adeptly handle the complexity and variability inherent in remote sensing data.

Machine learning methodologies offer data-driven solutions for mapping target features within remote sensing data and have witnessed substantial advancements in recent years \citep{kotsiantis2006machine,shirmard2022review,tawade2022remote}. Capable of handling high-dimensional data for a wide range of tasks, including compression, prediction, and classification \citep{bishop2006pattern,kotsiantis2006machine,lee2014proximal}, methods such as k-nearest neighbour (KNN), support vector machine (SVM), and multilayer perceptron (MLP) have been widely applied in Earth and environmental sciences. Their applications span diverse fields such as climate modelling \citep[e.g.,][]{chantry2021opportunities}, natural disaster prediction \citep[e.g.,][]{linardos2022machine}, environmental impact assessment \citep[e.g.,][]{tahmasebi2020machine}, and mineral exploration \citep[e.g.,][]{shirmard2022review}. Their particular strength is handling noisy, sparse, complex, and high-dimensional features \citep{zhang2002association}. Advanced machine learning techniques, such as deep learning models have driven a paradigm shift in remote sensing data analysis. These methodologies have facilitated the development of precise and efficient approaches for classifying remote sensing data and mapping geological features, encompassing lithological units, structural features, and alteration zones \citep{shirmard2022review}.

Deep learning models, such as convolutional neural networks (CNNs) \citep[e.g.,][]{chen2018recognition,shakya2021parametric}, recurrent neural networks (RNNs) \citep[e.g.,][]{ienco2017land,liu2022building}, and deep belief networks (DBNs) \citep[e.g.,][]{zhong2016diversified,li2019deep}, have gained prominence in automatically extracting features from remote sensing data. CNNs have demonstrated their utility in classifying various land uses and land covers \citep{ma2019evaluation}, while RNNs have proven effective in classifying time series data, including satellite imagery \citep{pelletier2019deep}. Applying deep learning methods becomes particularly valuable in alteration mapping, as they can discern specific mineralogical changes arising from mineralisation processes. These changes, often challenging to identify through visual interpretation and traditional methods \citep{he2010hydrothermal,bhadra2013aster,farahbakhsh2016fusing}, can be accurately pinpointed through the extraction of subtle features from remote sensing data using deep learning methods. This capability enhances the precision of alteration zone identification, marking a significant advancement in the field.

CNNs are prominent for their proficiency in automatic feature extraction in image-based data \citep{alzubaidi2021review}. Their versatility in processing multidimensional data arrays makes them well-suited for handling multispectral data characterised by evenly spaced pixels \citep{ma2019deep}. Comprising a stack of convolution layers and filters, CNNs excel in extracting hierarchical features, obviating the need for separate feature extraction in image pre-processing \citep{maggiori2016convolutional}. The efficacy of CNNs extends across various domains, spanning computer vision, natural language processing, and speech recognition \citep{kavitha2014speech,karpathy2015deep,li2022stm}. These achievements have spurred a growing interest in leveraging CNNs for processing remote sensing data, owing to their unique ability to extract intricate spatial features from extensive datasets \citep{song2019survey,wu2021convolutional}. Designed to discern patterns in images by learning pixel relationships \citep{maggiori2016convolutional}, CNNs are adept at classifying and segmenting large volumes of remote sensing data, thereby facilitating the creation of detailed and accurate maps \citep{maggiori2017convolutional}. CNNs are invaluable in identifying specific mineralogical changes arising from mineralisation processes. These changes, often elusive using traditional methods, become discernible as CNNs extract subtle features from remote sensing data, enabling the precise identification of alteration zones with exceptional accuracy. The rapid and accurate analysis of substantial datasets by CNNs empowers exploration geologists to pinpoint areas harbouring the highest potential for mineral deposits. Consequently, CNNs have emerged as a promising tool for processing remote sensing data and generating alteration maps in geological exploration.

In this study, we employ CNNs with remote sensing data to map diverse alteration types and optimise exploration efforts in the mineral-rich region north of Broken Hill in the far west of New South Wales (NSW), Australia. We include CNNs in our framework for alteration mapping that implements a classification approach using different types of remote sensing data. We also perform a comparative analysis within our framework using three conventional machine learning methods: KNN, SVM, and MLP. Our investigation leverages three types of multispectral remote sensing data, including Landsat 8, Landsat 9, and ASTER, for the study area. Additionally, we employ two sets of training samples. One originates from a restricted dataset comprising rock samples and geological maps from the study area, while the other is generated through selective principal component analysis, following the methodology outlined by \cite{shirmard2020integration}. In our framework, we evaluate the efficacy of selected models and data types in discriminating between different alteration types and generating a reliable alteration map through various performance metrics.

\section{Geological setting}

\begin{figure}
\centering
\includegraphics[width=\columnwidth]{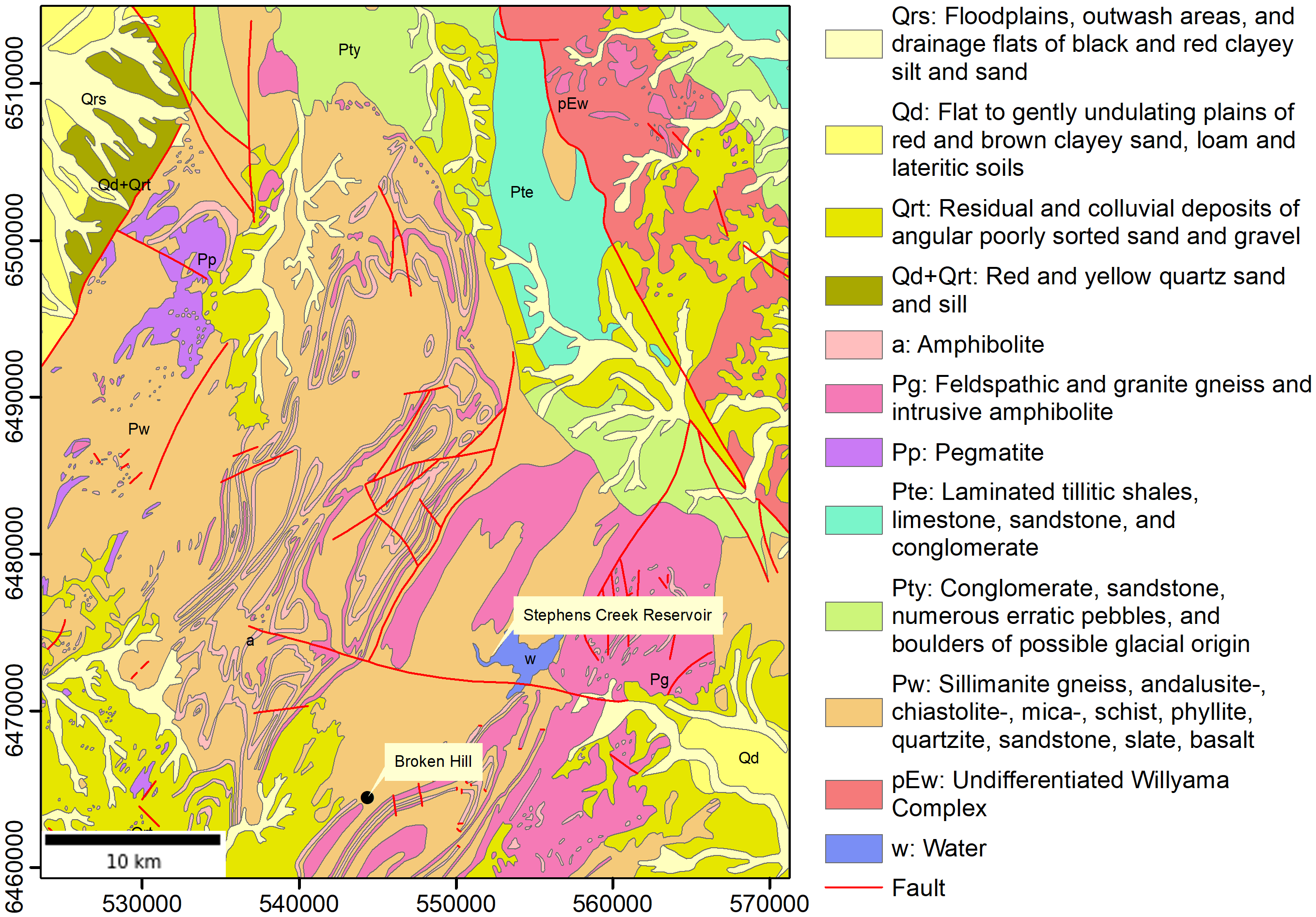}
\caption{Simplified geological map of the study area located in the far west of NSW, Australia. The coordinate system is WGS 84 / UTM zone 54S.}
\label{fig_01}
\end{figure}

The study area is located in the far west of New South Wales, Australia. It covers a vast area in the north and northeast of Broken Hill within the broader context of the Curnamona Province, which is part of the eastern margin of the Gawler Craton. The region has a complex geological history involving multiple tectonic events, resulting in diverse rock types and structures. The oldest rocks in the region are part of the Willyama Supergroup, which consists of sedimentary rocks ranging in age from the Neoproterozoic to the Cambrian (Fig. \ref{fig_01}) \citep{stevens1988willyama}. These rocks are overlain by a sequence of Ordovician to Silurian sediments, including the Broken Hill Group. The Broken Hill Group is known for its rich lead, zinc, and silver deposits, mined extensively in the area \citep{morland1998broken}. In addition to sedimentary rocks, the region is also characterised by intrusive and extrusive igneous rocks. The Devonian to Carboniferous Burra Group is a suite of mafic and ultramafic rocks that intrude the sedimentary rocks of the Curnamona Province. The Burra Group is believed to have been formed by the ascent of partial melts from the mantle wedge above a subduction zone \citep{ireland1998development}. The Curnamona Province has also been affected by a number of tectonic events, including the Delamerian Orogeny, which occurred during the Late Ordovician to Early Devonian, and the Petermann Orogeny, which occurred during the Late Neoproterozoic to Early Cambrian \citep{fergusson2015early}. These events have resulted in the deformation and folding of the rocks in the region, as well as the development of faults and fractures.

The mineralisation potential in Broken Hill is primarily associated with several types of mineral deposits, including stratabound, vein-hosted, and disseminated deposits \citep{morland1998broken}. Stratabound deposits are the most common type of mineralisation in the area and are typically associated with sedimentary rocks that have undergone hydrothermal alteration. These deposits are characterised by the presence of massive sulphide ores, which can contain high concentrations of base metals such as lead, zinc, and copper \citep{plimer1978proximal}. Vein-hosted deposits are another important type of mineralisation in the region and are often associated with faults and fractures filled with mineral-rich fluids. These deposits are typically narrow, high-grade, and contain a variety of precious and base metals \citep{large2005stratiform}. Disseminated deposits are less common in the area, but are still present. These deposits are characterised by low-grade mineralisation distributed throughout a rock formation rather than concentrated in a single deposit \citep{parr2004subseafloor}. The mineralisation potential in the study area is significant, offering a diverse range of mineralisation types and deposit styles. The geology, alteration zones, and mineral assemblages of the study area provide valuable clues for effective mineral exploration and resource assessment.

Broken Hill is known for extensive alteration zones indicative of mineralisation processes. These alteration zones are of great interest to geologists seeking to identify potential mineral deposits. The alteration zones are primarily associated with the Broken Hill Group, and the most common alteration types in the region include silicification, propylitic, argillic, and hematisation \citep{spry2021classification}. Silicification is characterised by replacing original minerals with silica \citep{spry2021classification}, often associated with mineralisation, and can indicate the presence of precious and base metals. Propylitic alteration is another common alteration type in the area, characterised by iron and magnesium-bearing hydrothermal fluids altering biotite or amphibole to minerals such as chlorite, epidote, and sericite. This type of alteration can indicate the presence of copper and other base metals \citep{pacey2016propylitic}. Argillic alteration is a type of hydrothermal alteration commonly found in the area and can be characterised by the replacement of plagioclase and amphibole with clay minerals, such as kaolinite, illite, and smectite. Argillic alteration is often associated with the upper parts of mineralising systems and can indicate the presence of gold, silver, and base metal deposits \citep{sillitoe1998advanced}. Hematisation is the other common alteration type in the area and is characterised by the replacement of ferromagnesian silicate minerals with hematite. This type of alteration is often associated with the oxidation of sulphide minerals and can indicate the presence of iron oxide-copper-gold (IOCG) deposits \citep{williams2005iron}.

\section{Materials and methods}
\subsection{Remote sensing data and pre-processing}

We implement our framework to map various alteration types using three types of multispectral remote sensing data with different characteristics, including Landsat 8, Landsat 9, and ASTER (advanced spaceborne thermal emission and reflection radiometer) satellite datasets. Landsat 8 was launched in 2013 and carries two sensors, the operational land imager (OLI) and thermal infrared sensor (TIRS). It captures images in 11 spectral bands, with spatial resolution ranging from 15 meters (m) for the panchromatic band to 30 m in the visible and near-infrared (VNIR) and short-wave infrared (SWIR) ranges, except for the last two thermal bands, 10 and 11, which have a resolution of 100 m \citep{roy2014landsat}. Landsat 9 was launched in 2021 and is similar to Landsat 8 in some ways, but it also has some notable differences, including its sensors and capabilities. Landsat 9 features an improved version of OLI called OLI-2 and an advanced TIRS (ATIRS) with improved signal-to-noise ratios and higher radiometric resolution, leading to better data accuracy and quality \citep{niroumand2022river}. In 1999, the ASTER sensor was launched on the Terra platform, greatly enhancing the capacity of geological remote sensing for mapping applications. ASTER features three VNIR bands with a 15 m spatial resolution, six SWIR bands with a 30 m resolution, and five thermal infrared bands with a 90 m resolution \citep{abrams2000advanced}.

In our study, we incorporate spectral bands crucial in geological remote sensing, leveraging their distinctive behaviours—such as high reflectance or absorption in various geological units—to enhance the production of high-quality alteration maps. In this context, we specifically select seven OLI/OLI-2 bands (bands 1 to 7) and nine ASTER bands (bands 1 to 9) as input for our framework. The absence of the blue band in ASTER data presents challenges for mapping iron oxide areas, making Landsat data more appropriate for this type of alteration. Conversely, ASTER's superior spectral resolution in the SWIR region makes it effective for mapping propylitic alteration zones. Both data types, however, are suitable for mapping argillic alteration type, as discussed in more detail by \cite{shirmard2020integration}. Therefore, we focus on delineating argillic and iron oxide areas using Landsat data by meticulously analysing the spectral features indicative of the target alteration zones and considering the spectral resolution. Simultaneously, we map argillic and propylitic alteration zones using ASTER data, aligning with their association to the formation of precious and base metal deposits in the study area \citep{pour2015hydrothermal,farahbakhsh2016fusing,shirmard2020integration}.

Cloud-free Landsat 8 and Landsat 9 images covering the study area are acquired from the Earth Resources Observation and Science Center (EROS) of the US Geological Survey (USGS)\footnote{\url{https://earthexplorer.usgs.gov}}. These images, categorised as level-1T (terrain corrected), were captured on September 26, 2021, and March 29, 2022, respectively. The ASTER image utilised in this study dates back to January 8, 2002, and is a cloud-free level-1-precision terrain-corrected registered at-sensor radiance product (ASTER\_L1T) sourced from the USGS EROS centre. All remote sensing datasets employed in this study are pre-georeferenced to the universal transverse Mercator (UTM) zone 54 South, obviating the need for geometric correction. Furthermore, the datasets undergo radiometric correction using the QUick Atmospheric Correction (QUAC) module within the ENVI software package \citep{harris2022envi}, with the resultant reflectance data serving as input for the machine learning algorithms.

The QUAC module incorporates various inputs, including atmospheric pressure, water vapour content, and aerosol optical depth, to execute atmospheric correction \citep{bernstein2012quick}. The module employs essential information about sensor viewing geometry, encompassing solar and viewing angles. The QUAC module effectively eliminates atmospheric effects from the data \citep{bernstein2012quick} by calculating atmospheric path radiance and subtracting it from the measured radiance. In order to maintain consistency, we resample the SWIR bands of the ASTER data to a 15 m spatial resolution, aligning with the VNIR bands, using the nearest neighbour method. We create a combined data layer encompassing the VNIR and SWIR bands and resize all the images to align with the dimensions of the target area for further processing.

\subsection{Model training datasets}

Our framework uses a supervised machine learning-based classification approach, which requires a labelled dataset. In remote sensing applications, the availability of labelled data for training machine learning models is often limited. Creating a labelled or training dataset in geological studies poses unique challenges, such as inherent variability, subjectivity, cost and time constraints along with data quality issues associated with geological datasets. Tackling these challenges necessitates meticulous planning, expertise, and collaborative efforts involving geologists, data scientists, and other domain experts. Our study addresses these complexities by utilising two distinct training datasets, each generated through different approaches.

\begin{figure*}
\centering
\includegraphics[width=0.7\textwidth]{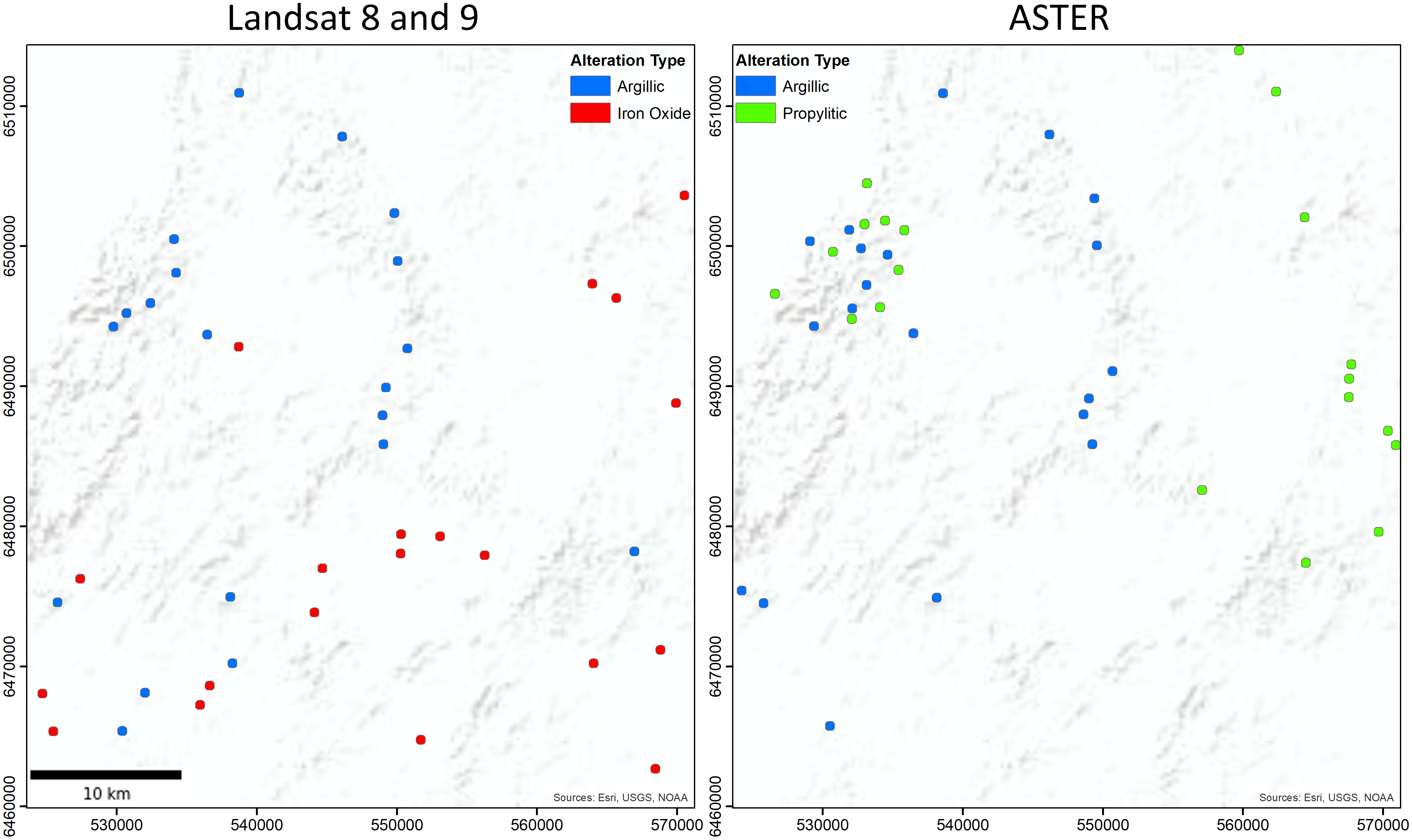}
\caption{Training samples generated using ground truth data for Landsat 8, Landsat 9, and ASTER data.}
\label{fig_02}
\end{figure*}

We create the first dataset by leveraging maps and reports based on ground truth data provided by the Geological Survey of New South Wales \footnote{\url{https://minview.geoscience.nsw.gov.au}} that features alteration zones in the study area. This involves the manual delineation of polygons and the subsequent labelling of pixels falling within these polygons (Fig. \ref{fig_02}). Notably, the set of training samples for Landsat 8 and Landsat 9 remains consistent, given their identical spatial resolution and the matching number of pixels along the x and y axes.

\begin{figure*}
\centering
\includegraphics[width=\textwidth]{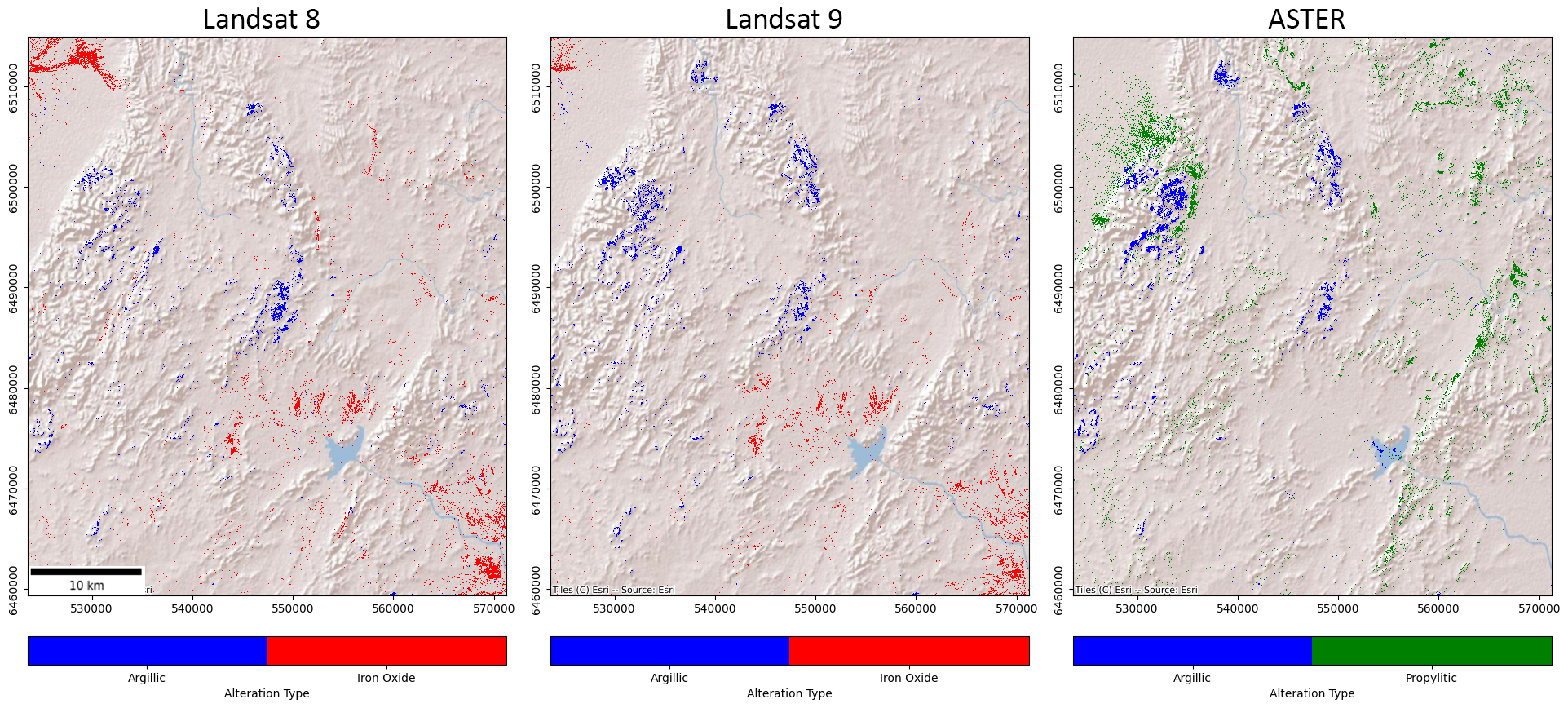}
\caption{Training samples generated by applying selective PCA \citep{shirmard2020integration} to Landsat 8, Landsat 9, and ASTER data.}
\label{fig_03}
\end{figure*}

The second set of training data adopts an approach presented by \cite{shirmard2020integration} that employs dimensionality reduction methods, featuring principal component analysis (PCA) \cite{jolliffe2016principal} to extract pixels highly indicative of target alteration zones, including those indicative of argillic, propylitic, and iron-oxide enrichment. The selection of appropriate principal components relative to these zones hinges on eigenvector loadings, indicating the relative strength of correlation between a given variable and a particular principal component. We select the key principal components based on eigenvector loadings that exhibit trends similar to the spectral characteristics of indicator minerals. Specifically, the selected components manifest unique contributions in both magnitude and sign for the absorption and reflection bands of specific alteration minerals or mineral groups. This ensures that the principal component-based image selected to enhance a particular alteration type exhibits distinct eigenvector loadings corresponding to the spectral characteristics of the target minerals \citep{shirmard2020integration}. Fig. \ref{fig_03} visually illustrates the second set of training samples generated using PCA for various remote sensing data and alteration types. Table \ref{table_1} details the total number of training samples, defined by pixels, for each alteration and data type. Each sample (pixel) includes the spectral data captured by individual bands (represented as vectors), providing detailed information across a range of wavelengths.

\begin{table*}
\centering
\caption{Total number of training samples (pixels) generated using different methods for each alteration and data type.}
\label{table_1}
\begin{tabular}{|c|ccc|ccc|}
\hline
\multirow{2}{*}{Data Type/Method} & \multicolumn{3}{c|}{Manually Generated Training Dataset}                     & \multicolumn{3}{c|}{PCA-based Training Dataset}                              \\ \cline{2-7} 
                                  & \multicolumn{1}{c|}{Iron Oxide} & \multicolumn{1}{c|}{Argillic} & Propylitic & \multicolumn{1}{c|}{Iron Oxide} & \multicolumn{1}{c|}{Argillic} & Propylitic \\ \hline
Landsat 8                         & \multicolumn{1}{c|}{8000}       & \multicolumn{1}{c|}{8000}     & -          & \multicolumn{1}{c|}{44159}      & \multicolumn{1}{c|}{20373}    & -          \\ \hline
Landsat 9                         & \multicolumn{1}{c|}{8000}       & \multicolumn{1}{c|}{8000}     & -          & \multicolumn{1}{c|}{27180}      & \multicolumn{1}{c|}{27078}    & -          \\ \hline
ASTER                             & \multicolumn{1}{c|}{-}          & \multicolumn{1}{c|}{32000}    & 32000      & \multicolumn{1}{c|}{-}          & \multicolumn{1}{c|}{107676}   & 284627     \\ \hline
\end{tabular}
\end{table*}

\subsection{Convolutional neural networks}

CNNs model is a prominent deep learning model commonly used in image processing and computer vision tasks such as image classification, object detection, and image segmentation \citep{zhiqiang2017review,aamir2019optimized}. CNNs have been designed to automatically learn spatial hierarchies of features from raw input data, making them well-suited for image-based classification tasks. CNNs consist of multiple layers, each with a specific function. The three main types of layers in a CNN are convolutional layers, pooling layers, and fully connected layers \citep{aamir2019optimized}. Convolutional layers are the heart of a CNN and perform a series of convolutions on the input data. The convolution operation involves sliding a filter (also known as a kernel) over the input data and computing the dot product between the filter and a small section of the input data at each position. This produces a feature map that highlights the presence of specific features in the input data. Mathematically, the convolution operation can be expressed as \citep{aamir2019optimized}:

\begin{equation}
y_{i,j} = f(\sum_{m} \sum_{n} w_{m,n}x_{i+m,j+n} + b)
\end{equation}

where $y_{i,j}$ is the output feature map at position $(i,j)$, $x_{i+m,j+n}$ is the input data at position $(i+m,j+n)$, $w_{m,n}$ is the weight of the filter at position $(m,n)$, $b$ is the bias term, and $f$ is the activation function. Pooling layers reduce the spatial dimensionality of the feature maps by performing a down-sampling operation. The most common pooling operation is max pooling, which selects the maximum value from a small region of the feature map \citep{nagi2011maxpooling}. This operation helps to make the network more robust to small shifts and distortions in the input data. Fully connected layers take the flattened output of the previous layer and perform a linear transformation on it. The output of the final fully connected layer is fed into a softmax activation function, which produces a probability distribution over the different classes \citep{wang2018highspeed}. During training, CNNs learn the optimal values of the weights and biases for each layer by minimising a loss function, which measures the difference between the predicted output and the true label. This is typically done using stochastic gradient descent and backpropagation, which computes the gradients of the loss function with respect to the weights and biases and updates them accordingly.

\subsection{Framework for alteration mapping}

\begin{figure}
\centering
\includegraphics[width=\columnwidth]{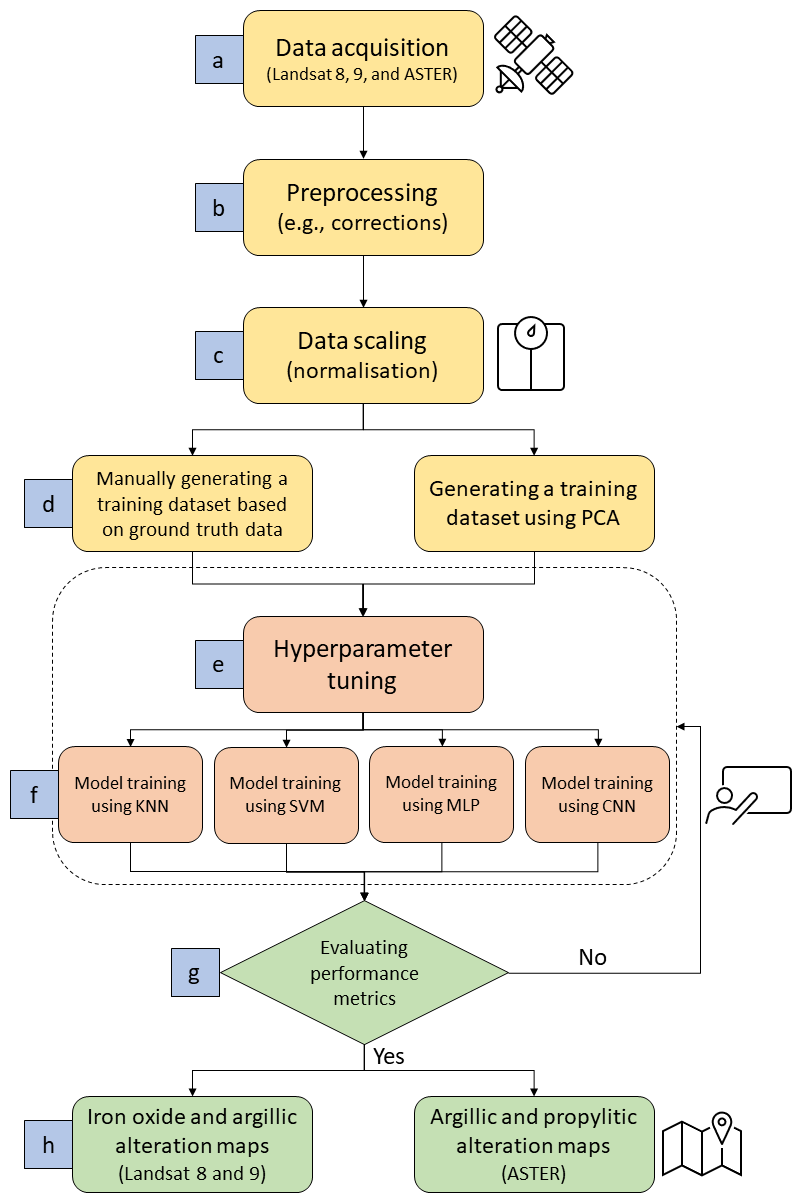}
\caption{Machine learning framework for mapping alteration zones utilising remote sensing data. The framework features KNN, SVM, MLP, and CNNs for the classification module to create alteration maps.}
\label{fig_04}
\end{figure}

Our machine learning framework for alteration mapping (Fig. \ref{fig_04}) features selected machine learning models for classification. We utilise three multispectral remote sensing datasets obtained from different satellites, including Landsat 8, Landsat 9, and ASTER (Fig. \ref{fig_04}-(a)). The Landsat 8 and 9 datasets exhibit dimensions of 1587 $\times$ 1854 pixels, while the ASTER image surpasses these dimensions with a size of 3175 $\times$ 3708 pixels. Each pixel within the Landsat and ASTER datasets corresponds to a square cell covering 900 and 225 square meters of the ground surface, respectively. The objective is to assess the efficacy of the selected machine learning models for mapping various alteration zones. The framework begins with acquiring, importing, and preparing remote sensing data. These steps include preprocessing (Fig. \ref{fig_04}-(b)), which includes any correction by removal of noise from data to enhance the clarity and usability of the data by reducing errors and distortions introduced by various sources, such as atmospheric conditions, sensor malfunctions, or environmental interference. It also includes downscaling the data to enhance the spatial resolution of coarser data to obtain finer, more detailed information.

We rescale the reflectance values of pixels to ensure that we assign uniform importance to all features in the models (Fig. \ref{fig_04}-(c)). This step is crucial since machine learning models can be sensitive to the input data distribution and may inadvertently assign more significance to features with higher values. Consequently, data scaling becomes necessary, either in the range of zero to one (normalised) or minus one to one (standardised). In trial experiments, we observed no substantial differences between the two methods; hence, we proceeded with data normalisation for training our models.

Next, our framework features the creation of training datasets that include two approaches (Fig. \ref{fig_04}-(d)). The first approach includes manually generating the training dataset using ground truth obtained from the information in the geological reports. The second approach includes employing dimensionality reduction methods (PCA) proposed by \cite{shirmard2020integration}. Subsequently, we allocate 70\% of the training datasets for model training and the remaining 30\% for testing, ensuring effective model evaluation. Then, we conduct trial experiments and employ optimal hyperparameters (Fig. \ref{fig_04}-(e)) to establish the most accurate machine learning models for each data type (Fig. \ref{fig_04}-(f)). Finally, we evaluate the respective models (KNN, SVM, MLP, and CNN) using a combination of metrics, such as classification accuracy and F1 score (Fig. \ref{fig_04}-(g)). We also create detailed alteration maps for visual analysis and provide visualisations of these maps and results (Fig. \ref{fig_04}-(h)).

\subsubsection{Model implementation}

We set the number of nearest neighbours to five in the KNN, which determines the complexity of the model. We also set the weight function used in prediction to uniform, meaning all points in each neighbourhood are weighted equally, which can influence the decision boundary. The regularisation parameter for the SVM algorithm, crucial in balancing the trade-off between achieving low training error and low testing error, is set to one. We adopt the radial basis function (RBF) as the kernel function—a common choice due to its ability to model complex nonlinear decision boundaries \citep{scholkopf1997comparing}. We compute the gamma parameter (width of the kernel) of the RBF using the inverse of the number of features in the training data.

\begin{figure}
\centering
\includegraphics[width=\columnwidth]{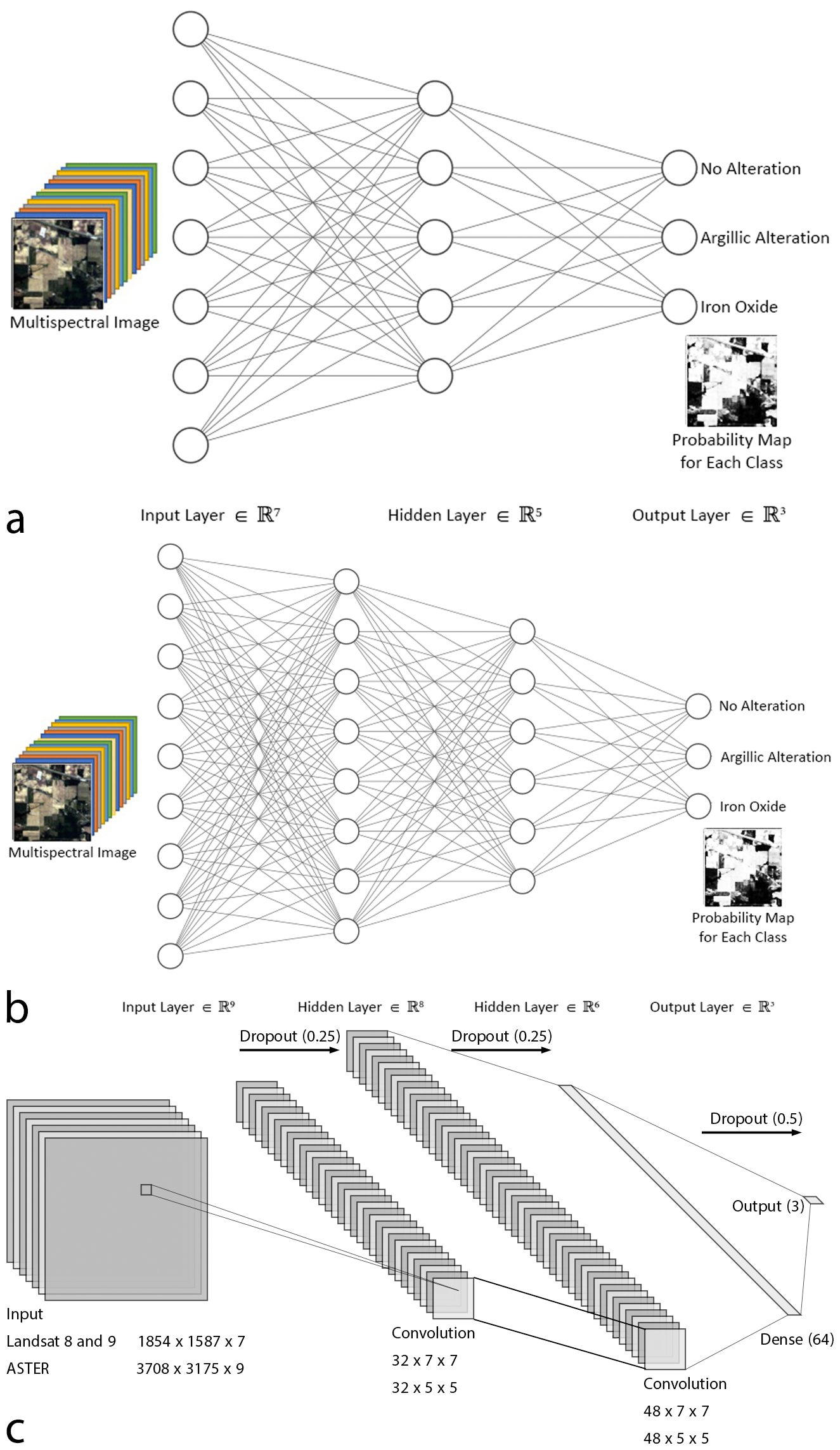}
\caption{Model architectures featuring a) MLP classifier for the Landsat data, b) MLP classifier for the ASTER data, and c) CNN classifier for mapping target alteration zones using both the Landsat and ASTER data.}
\label{fig_05}
\end{figure}

The architecture used for the MLP model varies between the Landsat and ASTER datasets due to the different number of spectral bands in each. Consequently, we designed distinct architectures to accommodate the specific characteristics and requirements of each dataset. Fig. \ref{fig_05}a presents the MLP architecture for the Landsat datasets, which employs one hidden layer featuring five neurons. The activation function in this layer is the scaled exponential linear unit (SELU), chosen for its ability to enhance model training performance \citep{klambauer2017self}. Subsequently, we utilise the softmax activation function in the output layer for generating a probability distribution over potential classes, an approach commonly employed in multi-class classification problems \citep{bishop2006pattern}. Fig. \ref{fig_05}b illustrates the MLP model architecture for the ASTER data, where two hidden layers feature eight and six neurons, respectively.

Fig. \ref{fig_05}c presents the CNN architecture for mapping target alteration zones across all datasets. The architectural configuration comprises two convolutional layers, incorporating 32 and 48 filters with a kernel size of seven and five for Landsat and ASTER data, respectively. We utilise the rectified linear unit (ReLU) activation function in these layers to account for non-linearity \citep{nair2010rectified}. We use a dropout-based regularisation layer to mitigate overfitting, \citep{srivastava2014dropout} with a rate of 0.25 in each convolutional layer. Finally, we use a fully connected layer with 64 neurons, and ReLU activation precedes a flattening of the output from convolutional layers into a one-dimensional vector. Finally, a dropout layer with a rate of 0.5 follows this fully connected layer, while the output layer features three neurons using the softmax activation function.

We train the respective MLP and CNN models using the Adam optimiser \citep{kingma2014adam} with a learning rate set at 0.01 and use the cross-entropy loss function to train the respective models for a different number of epochs depending on the data type and set of training samples. We use 50 and 100 epochs for the models in the case of the Landsat data using the PCA-based and ground truth-based training datasets, respectively. Similarly, ASTER data are trained for 20 and 40 epochs using the PCA-based and ground truth-based training datasets, respectively. Note that the respective model implementations employed a sequential model using the Keras Python deep learning library \footnote{\url{https://keras.io}} and our implementation is available via GitHub repository \footnote{\url{https://github.com/sydney-machine-learning/deeplearning_alteration_zones}}. 

\section{Results}

\begin{figure}
\centering
\includegraphics[width=\columnwidth]{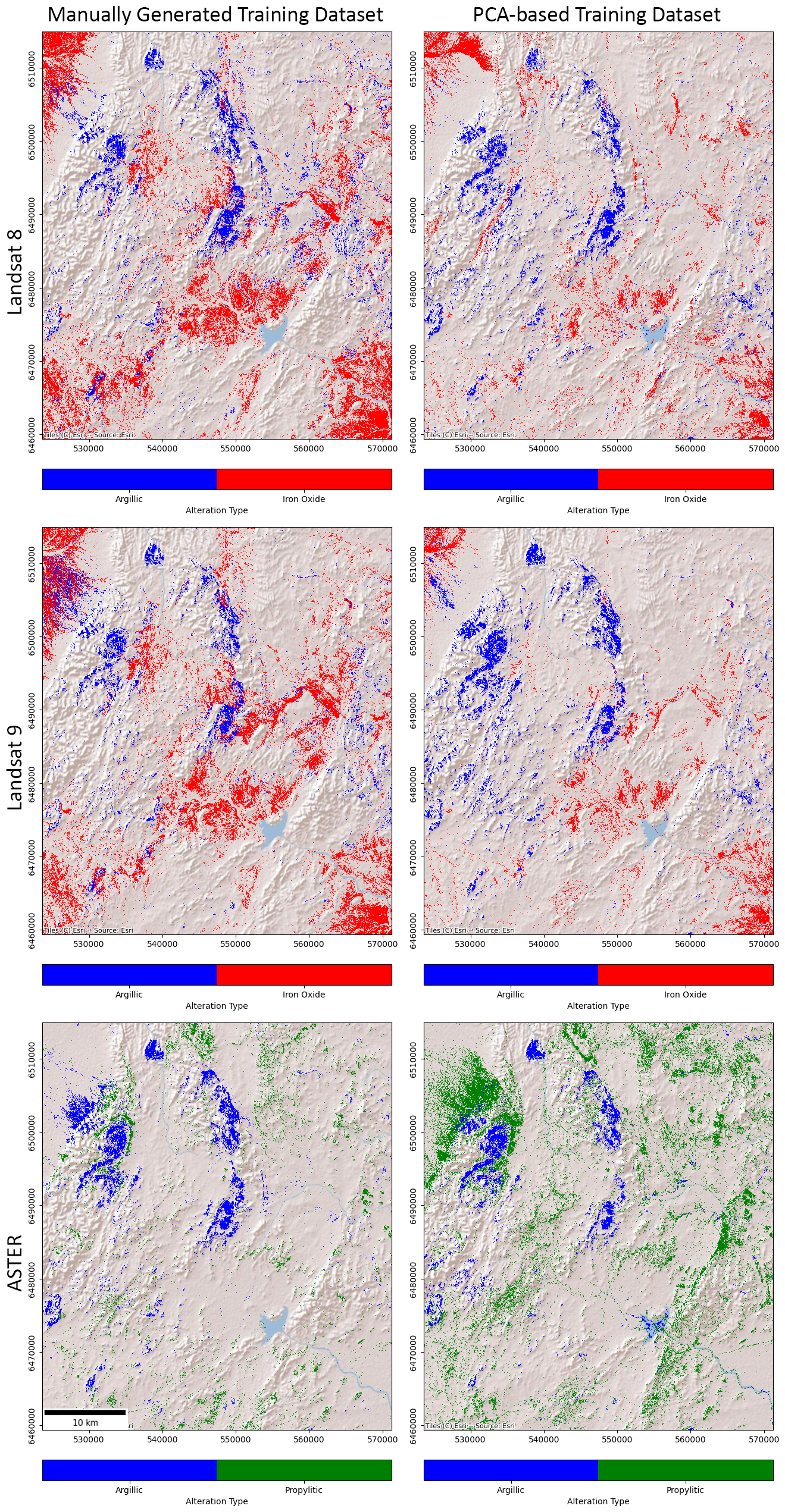}
\caption{Alteration maps produced through KNN utilising a combination of three distinct remote sensing datasets and two sets of training samples.}
\label{fig_06}
\end{figure}

\begin{figure}
\centering
\includegraphics[width=\columnwidth]{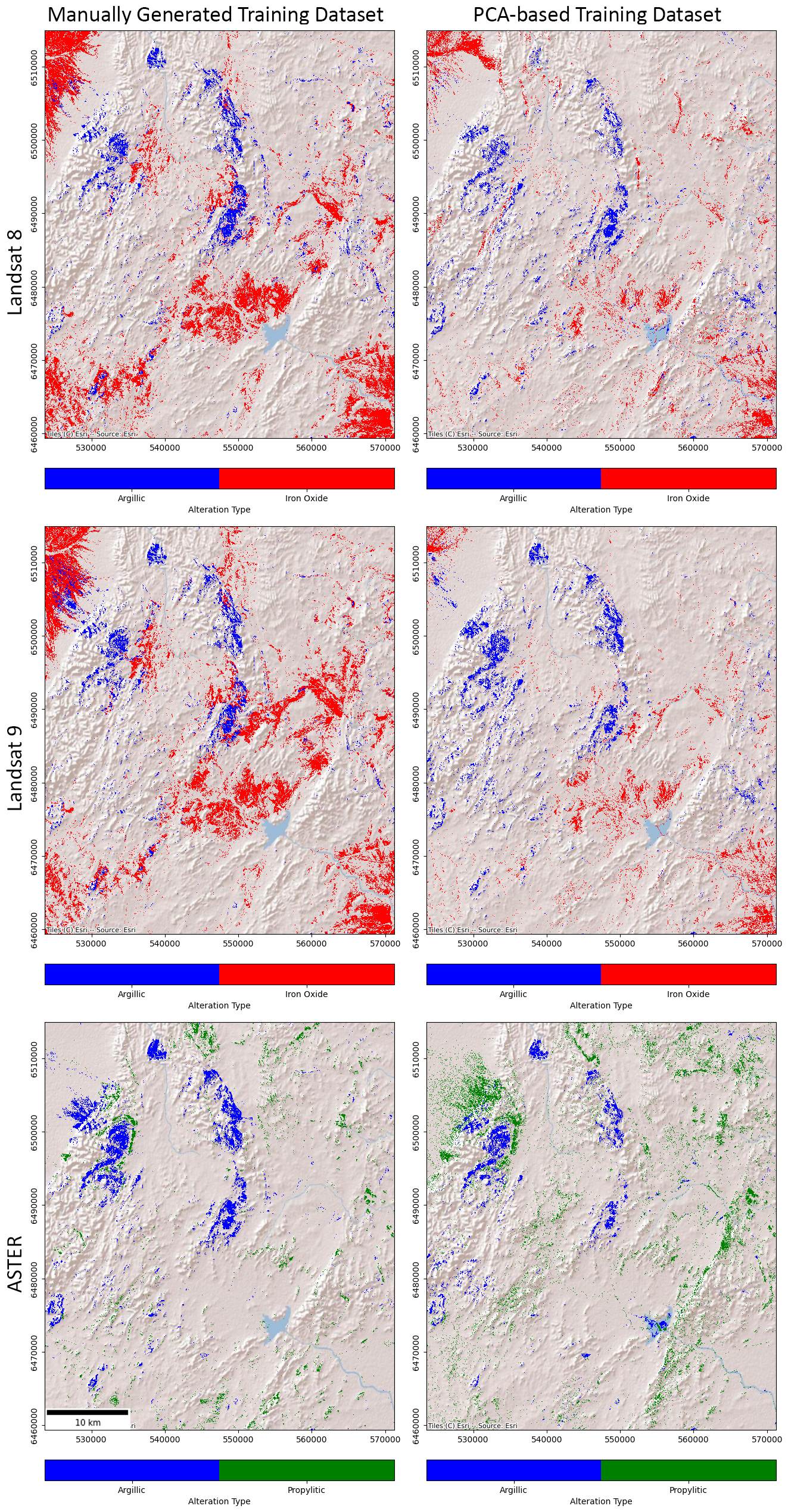}
\caption{Alteration maps produced through SVM utilising a combination of three distinct remote sensing datasets and two sets of training samples.}
\label{fig_07}
\end{figure}

\begin{figure}
\centering
\includegraphics[width=\columnwidth]{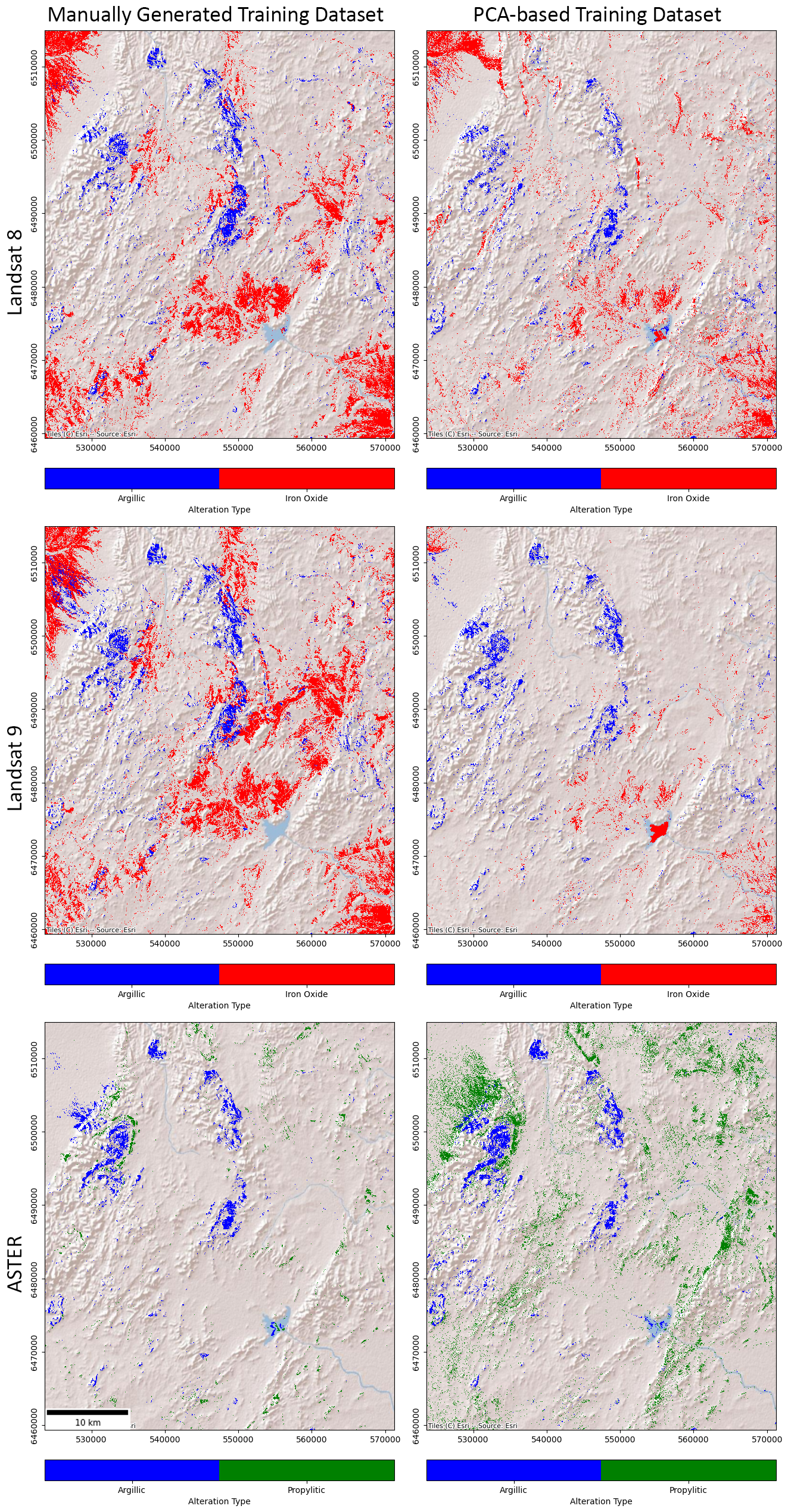}
\caption{Alteration maps produced through MLP utilising a combination of three distinct remote sensing datasets and two sets of training samples.}
\label{fig_08}
\end{figure}

\begin{figure}
\centering
\includegraphics[width=\columnwidth]{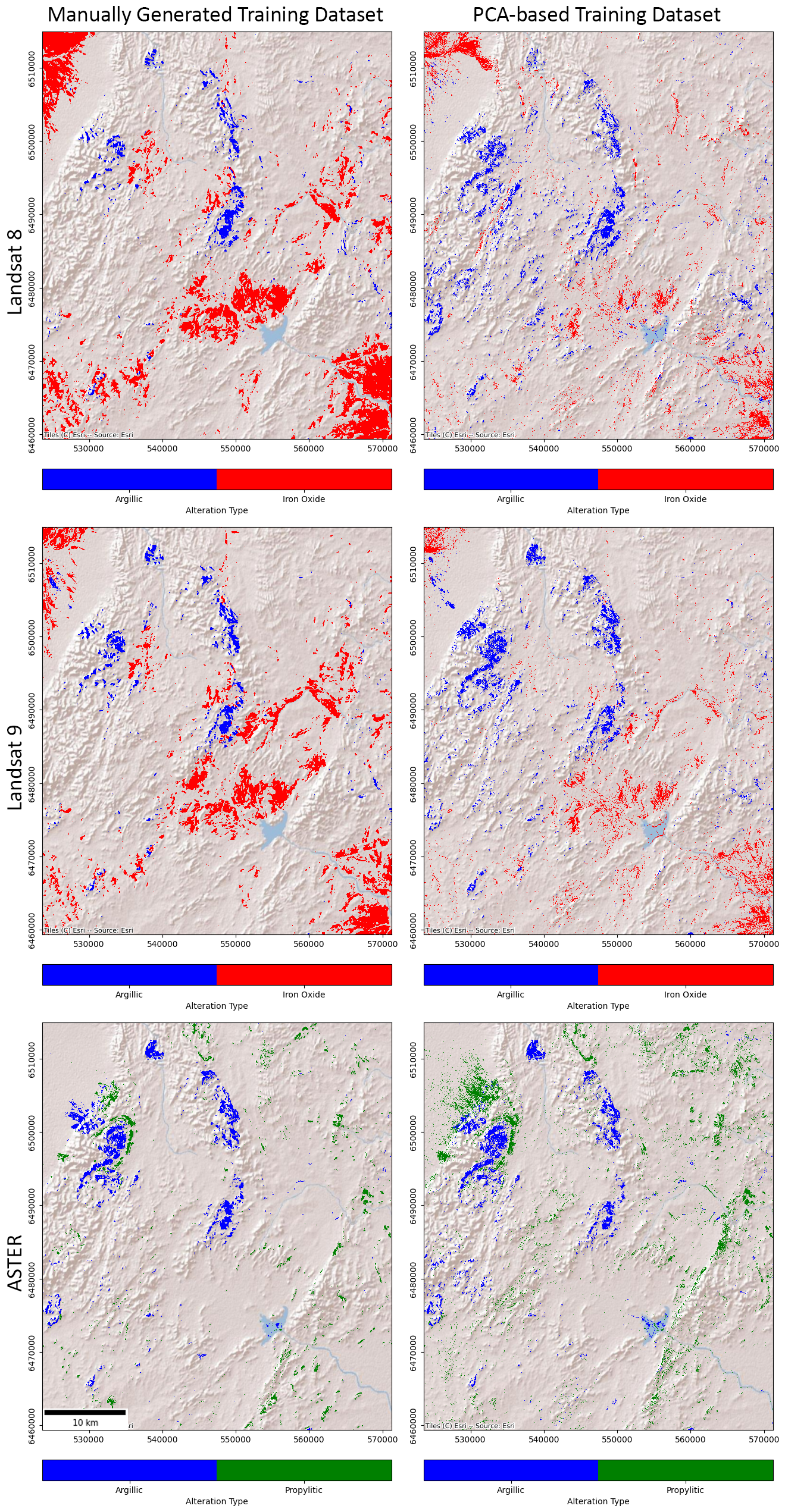}
\caption{Alteration maps produced through CNNs utilising a combination of three distinct remote sensing datasets and two sets of training samples.}
\label{fig_09}
\end{figure}

We employ our framework on specific spectral bands of three types of remote sensing data, selected based on the spectral features of the desired alteration zones in the study area. A distinctive alteration map is generated using each classifier, i.e., KNN, SVM, MLP, and CNN—utilising a pair of remote sensing data and training samples as depicted in Figs. \ref{fig_02} and Fig. \ref{fig_03}. In Figs. \ref{fig_06}--\ref{fig_09}, we present the resulting maps from the respective models. As mentioned earlier, we use the Landsat datasets to map argillic alteration and iron oxide areas, while ASTER data are utilised to map argillic and propylitic zones. Notably, the argillic alteration zones are concentrated in the central and western parts of the study area, while the propylitic zones and iron oxide areas are dispersed across the region. The spatial resolution of the maps created using the Landsat and ASTER datasets is 30 and 15 m, respectively, contingent on the spatial resolution of the input data. We report the performance metrics for each model, encompassing accuracy and F1 score, in Tables \ref{table_2}--\ref{table_5} for the respective models. We also assess the models produced using the PCA-based training samples by incorporating the ground truth data, presented in the last column of Tables \ref{table_2}--\ref{table_5}. This can be considered a strategy to mark the ratio of correctly predicted samples or pixels to the total number of pixels within the ground truth-based dataset. Utilising this strategy provides a more insightful measure for comparing the performance of the models.

The results in Tables \ref{table_2}--\ref{table_5} show that CNNs generally outperform conventional models when using manually generated training datasets. The improvement is higher for the Landsat 8 and 9 datasets when compared to ASTER. In the case of the PCA-based training samples with Landsat 8, CNNs greatly outperform conventional methods, achieving the best ground truth-based accuracy. However, the performance of SVM and MLP are close to CNNs when considering the highest overall accuracy and F1 score. In Landsat 9, the KNN model attains the highest ground truth-based accuracy, while the MLP model surpasses the others in terms of accuracy and F1 score. In the case of ASTER data, the KNN model again shows the highest ground truth-based accuracy, and the SVM model outperforms the rest based on accuracy and F1 score. Note that when using the PCA-based training dataset, the performance metrics are very close across the respective models in most of the cases, with negligible differences, resulting in similar performance across the models.

In the case of Landsat data, Landsat 9 with CNNs trained on manually generated data proves to be slightly more reliable for mapping alteration zones. This improvement is likely due to the higher radiometric resolution of Landsat 9 compared to Landsat 8. Similarly, with the manually generated training dataset, ASTER and the CNNs model offer the most accurate mapping of alteration zones. We observe that using PCA to generate the training datasets significantly increases the number of training samples, leading to higher overall accuracy and F1 scores when compared to the models created using ground truth-based samples. The exception is the CNNs model with ASTER data, where the performance metrics are lower than the model created using the ground truth-based training dataset. This is likely because CNNs are more sensitive to noisy data compared to conventional machine learning models, thanks to their convolutional and pooling layers, which enable hierarchical feature extraction. This process can amplify noise at deeper layers, potentially resulting in misclassifications or degraded performance. Using PCA-based training datasets to generate alteration maps offers a fully automated process. This process is free from reliance on ground truth data, which proves particularly advantageous in scenarios where the study area access is limited or a paucity of rock samples exists.

Furthermore, we find that the accuracy metric derived from ground truth data is more reliable than those based solely on PCA-based samples. As indicated in Tables \ref{table_2}--\ref{table_5}, the accuracy achieved with ground truth-based training samples surpasses that of PCA-based samples for the CNNs model with Landsat 9 data. However, for the KNN model with ASTER data, PCA-based training samples yield the highest accuracy. Additionally, the accuracy achieved with the combination of Landsat 8 and CNNs is the best, with no significant difference between the two training datasets. Overall, alteration maps generated with ground truth-based training samples exhibit greater interpretability and less noise.

\begin{table*}[]
\centering
\caption{Performance metrics of the models produced using KNN to map alteration zones.}
\label{table_2}
\begin{tabular}{|c|cc|ccc|}
\hline
\multirow{2}{*}{Data Type/Method} & \multicolumn{2}{c|}{Manually Generated Training Dataset} & \multicolumn{3}{c|}{PCA-based Training Dataset}                                         \\ \cline{2-6} 
                                  & \multicolumn{1}{c|}{Accuracy}         & F1 Score         & \multicolumn{1}{c|}{Accuracy} & \multicolumn{1}{c|}{F1 Score} & Accuracy (Ground Truth) \\ \hline
Landsat 8                         & \multicolumn{1}{c|}{0.913}            & 0.917            & \multicolumn{1}{c|}{0.974}    & \multicolumn{1}{c|}{0.973}    & 0.909                   \\ \hline
Landsat 9                         & \multicolumn{1}{c|}{0.929}            & 0.931            & \multicolumn{1}{c|}{0.976}    & \multicolumn{1}{c|}{0.976}    & 0.880                   \\ \hline
ASTER                             & \multicolumn{1}{c|}{0.974}            & 0.974            & \multicolumn{1}{c|}{0.959}    & \multicolumn{1}{c|}{0.963}    & 0.995                   \\ \hline
\end{tabular}
\end{table*}

\begin{table*}[]
\centering
\caption{Performance metrics of the models produced using SVM to map alteration zones.}
\label{table_3}
\begin{tabular}{|c|cc|ccc|}
\hline
\multirow{2}{*}{Data Type/Method} & \multicolumn{2}{c|}{Manually Generated Training Dataset} & \multicolumn{3}{c|}{PCA-based Training Dataset}                                         \\ \cline{2-6} 
                                  & \multicolumn{1}{c|}{Accuracy}         & F1 Score         & \multicolumn{1}{c|}{Accuracy} & \multicolumn{1}{c|}{F1 Score} & Accuracy (Ground Truth) \\ \hline
Landsat 8                         & \multicolumn{1}{c|}{0.949}            & 0.948            & \multicolumn{1}{c|}{0.987}    & \multicolumn{1}{c|}{0.986}    & 0.880                   \\ \hline
Landsat 9                         & \multicolumn{1}{c|}{0.898}            & 0.898            & \multicolumn{1}{c|}{0.987}    & \multicolumn{1}{c|}{0.987}    & 0.849                   \\ \hline
ASTER                             & \multicolumn{1}{c|}{0.974}            & 0.973            & \multicolumn{1}{c|}{0.988}    & \multicolumn{1}{c|}{0.988}    & 0.963                   \\ \hline
\end{tabular}
\end{table*}

\begin{table*}[]
\centering
\caption{Performance metrics of the models produced using MLP to map alteration zones.}
\label{table_4}
\begin{tabular}{|c|cc|ccc|}
\hline
\multirow{2}{*}{Data Type/Method} & \multicolumn{2}{c|}{Manually Generated Training Dataset} & \multicolumn{3}{c|}{PCA-based Training Dataset}                                         \\ \cline{2-6} 
                                  & \multicolumn{1}{c|}{Accuracy}         & F1 Score         & \multicolumn{1}{c|}{Accuracy} & \multicolumn{1}{c|}{F1 Score} & Accuracy (Ground Truth) \\ \hline
Landsat 8                         & \multicolumn{1}{c|}{0.936}            & 0.935            & \multicolumn{1}{c|}{0.983}    & \multicolumn{1}{c|}{0.984}    & 0.808                   \\ \hline
Landsat 9                         & \multicolumn{1}{c|}{0.940}            & 0.939            & \multicolumn{1}{c|}{0.988}    & \multicolumn{1}{c|}{0.988}    & 0.833                   \\ \hline
ASTER                             & \multicolumn{1}{c|}{0.977}            & 0.977            & \multicolumn{1}{c|}{0.982}    & \multicolumn{1}{c|}{0.984}    & 0.977                   \\ \hline
\end{tabular}
\end{table*}

\begin{table*}[]
\centering
\caption{Performance metrics of the models produced using CNNs to map alteration zones.}
\label{table_5}
\begin{tabular}{|c|cc|ccc|}
\hline
\multirow{2}{*}{Data Type/Method} & \multicolumn{2}{c|}{Manually Generated Training Dataset} & \multicolumn{3}{c|}{PCA-based Training Dataset}                                         \\ \cline{2-6} 
                                  & \multicolumn{1}{c|}{Accuracy}         & F1 Score         & \multicolumn{1}{c|}{Accuracy} & \multicolumn{1}{c|}{F1 Score} & Accuracy (Ground Truth) \\ \hline
Landsat 8                         & \multicolumn{1}{c|}{0.956}            & 0.957            & \multicolumn{1}{c|}{0.983}    & \multicolumn{1}{c|}{0.982}    & 0.956                   \\ \hline
Landsat 9                         & \multicolumn{1}{c|}{0.973}            & 0.974            & \multicolumn{1}{c|}{0.985}    & \multicolumn{1}{c|}{0.985}    & 0.868                   \\ \hline
ASTER                             & \multicolumn{1}{c|}{0.988}            & 0.988            & \multicolumn{1}{c|}{0.981}    & \multicolumn{1}{c|}{0.982}    & 0.975                   \\ \hline
\end{tabular}
\end{table*}

\begin{figure*}[htbp!]
\centering
\includegraphics[width=0.7\textwidth]{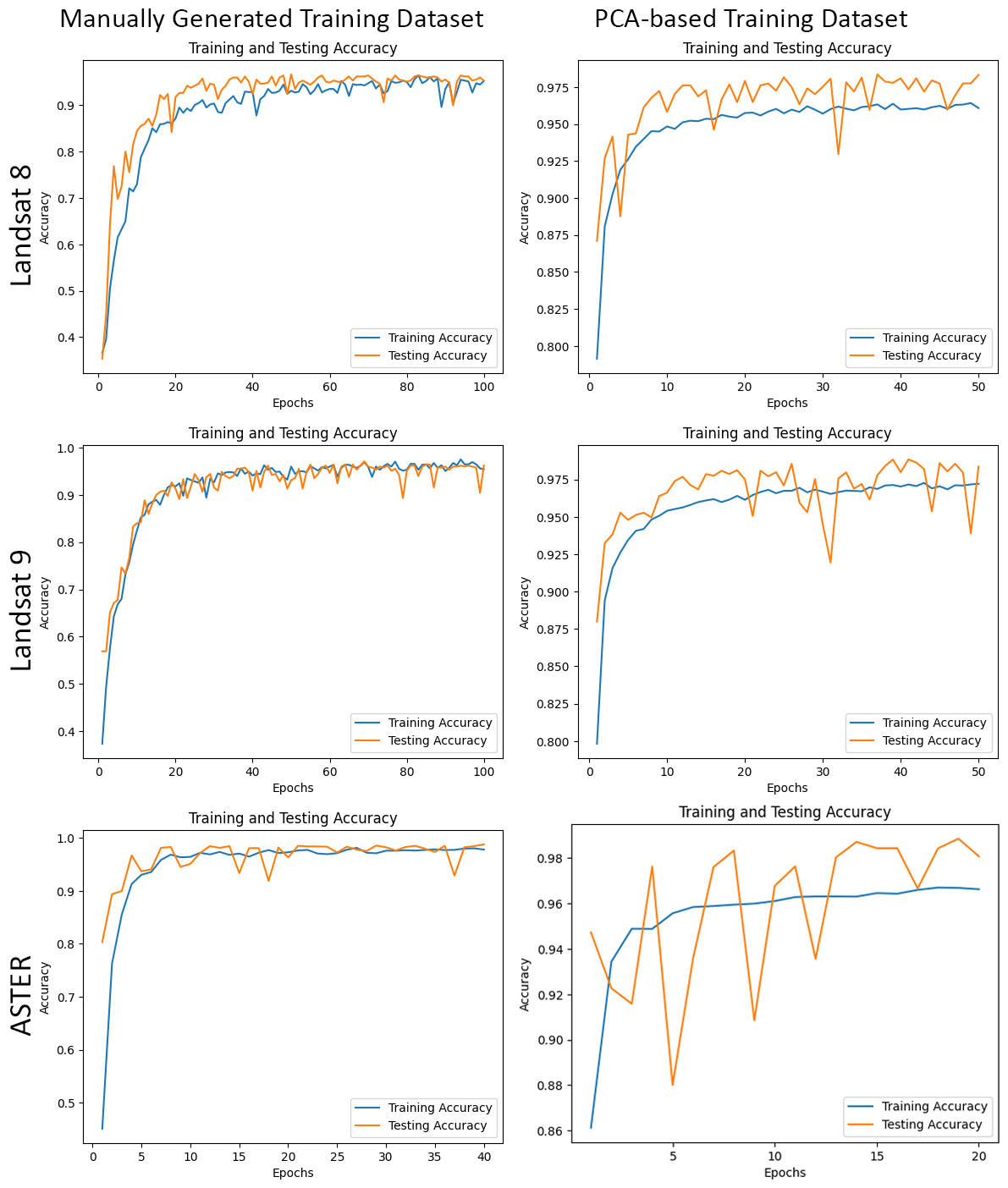}
\caption{Accuracy graphs of the models created using CNNs and different pairs of remote sensing and training datasets.}
\label{fig_10}
\end{figure*}

We present the training accuracy plot of the CNNs in Fig. \ref{fig_10} to review how the training and test accuracy evolve during the training process, measuring the disparity between predicted and actual outputs for a given input \citep{wang2020comprehensive}. We notice that the CNN models behave similarly across the different data types when looking at the manually generated training dataset. In this case, we notice that an acceptable training/test performance is achieved after 40 epochs. However, in the case of the PCA-based training dataset, we find that the trend is similar for the Landsat datasets but quite different when looking at the ASTER dataset, where more fluctuations are visible in the test accuracy over time. This could imply that Landsat datasets are more suitable for CNNs in the case of PCA-based training datasets.

\begin{figure*}
\centering
\includegraphics[width=0.7\textwidth]{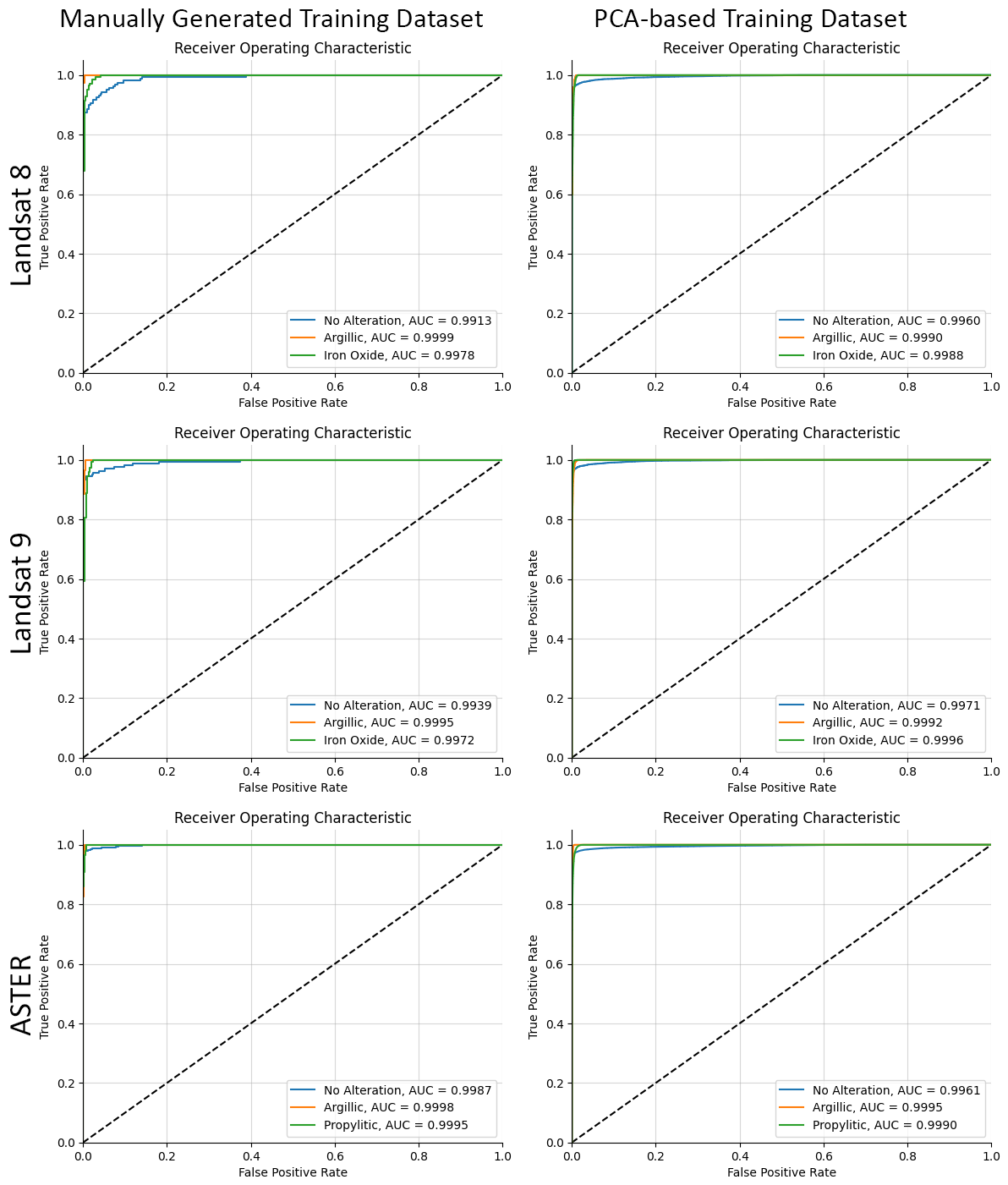}
\caption{ROC curves illustrating the performance of the models produced through CNN, utilising a combination of three distinct remote sensing datasets and two sets of training samples.}
\label{fig_11}
\end{figure*}

In addition to the performance metrics outlined in Tables \ref{table_2}--\ref{table_5}, we conduct an assessment of each model's efficiency and the pairing of remote sensing and training datasets through receiver operating characteristics (ROC) curve \citep{bradley1997use}. These curves portray the true-positive (sensitivity) rate on the y-axis against the false-positive rate (specificity) on the x-axis. The optimal point on the graph is situated at the top-left corner, where the false-positive rate is zero and the true-positive rate is one. Although the ideal scenario is often not realised in practice, a larger area under the curve (AUC) is generally favoured. The ROC curve typically lies between these extremes, with the AUC serving as a diagnostic accuracy measure summarising performance across various test values \citep{zweig1993receiver}. The steepness of the ROC curves is also relevant, emphasising the need to maximise the true-positive rate while minimising the false-positive rate. Fig. \ref{fig_11} presents the ROC curves of the CNN models to evaluate the performance using different combinations of data types with the AUC values. In the case of the conventional machine learning models, the ROC curves are available via Zenodo \footnote{\url{https://doi.org/10.5281/zenodo.11378720}}.

In Fig. \ref{fig_11}, we observe that the ROC curves reveal no significant disparity among various classifiers in mapping argillic alteration zones and iron oxide areas using Landsat data with the PCA-based training dataset. Furthermore, Landsat 8 and 9 generally demonstrate nearly identical results, albeit Landsat 9 yields slightly more accurate maps. We observe contrasting outcomes for alteration zones mapped with ASTER data when analysing Tables \ref{table_2}--\ref{table_5}, where SVM and MLP provide slightly better results when employing PCA-based training datasets. In a specific case, we found that CNNs exhibit non-converging loss function behaviour (Fig. \ref{fig_10}: testing accuracy for ASTER with PCA-based training data), possibly indicative of noise inherent in the training dataset, which could be due to the data type (ASTER). This underscores a limitation of CNNs when applied to large datasets with a substantial number of training samples. We note that the outcomes may vary in other locations, influenced by factors such as the size and geological properties of the area of interest, including the types of indicator minerals and rocks present.

\section{Discussion}

\begin{figure*}
\centering
\includegraphics[width=\textwidth]{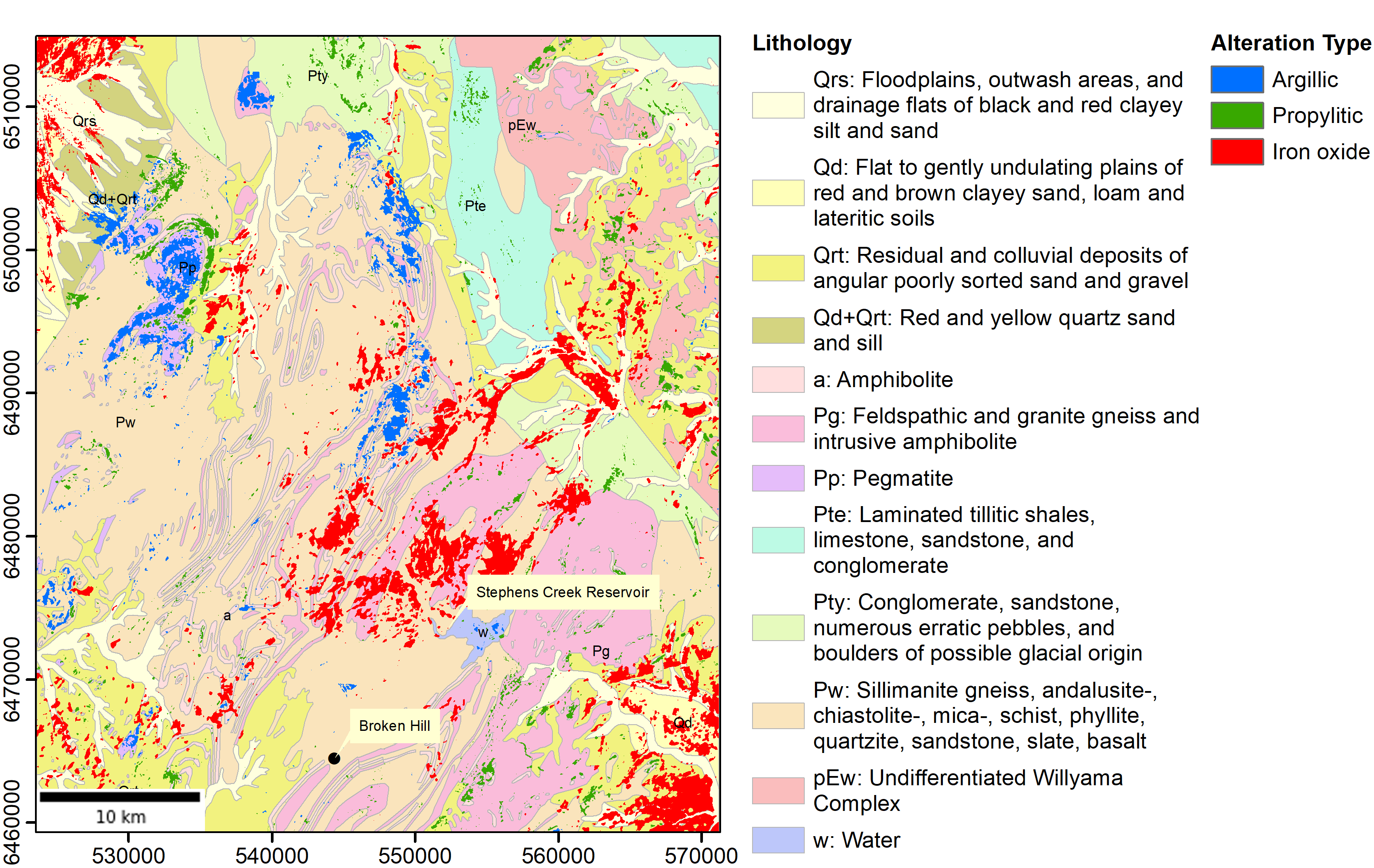}
\caption{Alteration zones, including argillic, propylitic, and iron oxide, mapped using CNNs, overlaid on the geological map of the study area.}
\label{fig_12}
\end{figure*}

In this study, we employed different machine learning models to classify remote sensing datasets, mapping areas of argillic, propylitic, and iron oxide, recognised as potential hydrothermal mineralisation zones for critical metals. To underscore the impact of training samples on alteration map accuracy, we utilised two distinct sets, i.e., manually generated and PCA-based training data. Argillic alteration zones, prominently observed in the western and central parts of the study area (Fig. \ref{fig_12}), show a close association with igneous units, specifically those housing sulphide minerals like pyrite \citep{sillitoe1998advanced}. These zones, characterised by clay minerals such as kaolinite and illite, signify weathering processes linked to these rock formations. The presence of argillic alteration, particularly in sulphide-bearing formations, is a crucial indicator for potential mineralisation. Propylitic alteration is concentrated in the northwest and east and is closely linked to intermediate to mafic volcanic rocks and less intense in the study area when compared to argillic alteration (Fig. \ref{fig_12}). This alteration results from hydrothermal fluid interaction with primary minerals in volcanic rocks, forming secondary minerals such as chlorite, epidote, and carbonates. The correlation between propylitic alteration zones and mineralisation is significant, often occurring in proximity to ore deposits containing base metals like lead and zinc \citep{morland1998broken}. Hematised or iron oxide-rich areas, prevalent in the northwestern, southeastern, and centre of the study area, involve the alteration of primary minerals to hematite, displaying a distinct reddish colour and a particular spectral behaviour (Fig. \ref{fig_12}). This alteration is commonly associated with sedimentary rocks containing iron-rich minerals \citep{large2005stratiform}. The noteworthy correlation between hematised zones and mineralisation indicates their significance as indicators, especially in proximity to orebodies associated with iron and other metals \citep{plimer1978proximal}.

The machine learning models used in this study offer unique strengths and limitations. KNN is simple and effective for small, well-separated datasets but struggles with high-dimensional data and is computationally intensive during prediction \citep{taunk2019brief}. In our study, due to the low number of input features, this method produced promising results, even yielding the most accurate alteration map when pairing Landsat 9 and ASTER data with the PCA-based training dataset. SVM excels in high-dimensional spaces with robust margin-based optimisation and effective kernel tricks for non-linear classification, but it can be inefficient for large datasets and requires careful parameter tuning \cite{noble2006support}. Our SVM models created with Landsat 8 and ASTER data were the most accurate and provided the highest F1 score when using the PCA-based training dataset. MLP is a simple neural network that can model complex nonlinear relationships and is suitable for various applications, though it demands significant computational resources and careful hyperparameter tuning to avoid overfitting \cite{arulampalam2003generalized}. Accordingly, we fine-tuned the MLP models and observed that the pair of Landsat 9 data with the PCA-based training dataset achieved the highest accuracy and F1 score.

CNNs have been powerful for image and spatial data analysis, efficiently capturing hierarchical features \citep{kreinovich2020deep}. However, they require large labelled datasets and substantial computational power, making them complex and resource-intensive. CNNs provide great flexibility and superior performance for more complex, large-scale data, especially in image recognition and spatial pattern analysis \cite{kattenborn2021review,alzubaidi2021review}. Our results show that when conventional methods are used appropriately and hyperparameter tuning is performed carefully, the difference in performance between CNNs and conventional methods can be significantly reduced. Our study underscores that CNNs, with their ability to extract features from images and handle spatial information \citep{li2021survey,rawat2017deep}, emerge as a valuable tool for mapping alteration zones, thereby contributing to the development of high-quality geological maps and advancing mineral exploration efforts.

\begin{figure*}
\centering
\includegraphics[width=\textwidth]{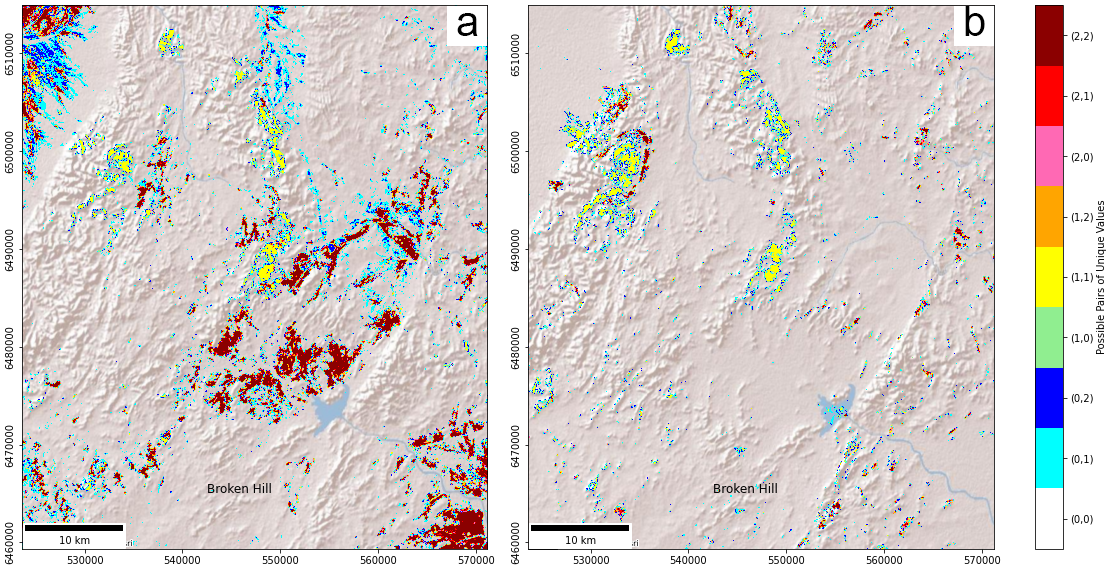}
\caption{Comparison of alteration zones mapped by CNNs and MLP using (a) Landsat 9 and (b) ASTER data. In the legend, the first element in parentheses indicates the alteration type identified by CNNs, while the second element represents the type mapped by MLP. For (a), 1 corresponds to argillic alteration and 2 to iron oxide areas. In (b), 1 denotes argillic alteration, and 2 indicates propylitic alteration.}
\label{fig_13}
\end{figure*}

In Fig. \ref{fig_13}, the alteration zones mapped using MLP, a conventional type of neural network, and CNNs, based on Landsat 9 and ASTER data, are compared. The color-coded legend illustrates the overlaps and discrepancies between the two models. The cyan and blue areas ((0,1) and (0,2)) indicate regions where only MLP detected alterations. The overall distribution patterns highlight variations in the models' sensitivity and accuracy. The MLP model captures broader alteration zones, especially in iron oxide areas identified by Landsat 9 (Fig. \ref{fig_13}a), while the ASTER data (Fig. \ref{fig_13}b) reveals more localised and distinct alteration zones. CNNs yield maps with reduced noise, attributed to their consideration of neighbouring pixels in mo1delling and reliance on patterns and textures rather than individual pixels. This comparison underscores the influence of sensor type and model architecture on the detection and classification of geological alterations.

The high accuracy observed in the CNN models (Table \ref{table_5}) can be attributed to the low number of mixed pixels within our testing set; however, it is crucial to acknowledge that more complex geological settings featuring mixed pixels could potentially lead to lower accuracy. The proficiency of the CNN models in predicting test classes stems from the easily distinguishable classes in the multidimensional space, facilitating the capacity to learn differentiation for pure pixels. The alteration maps generated by our framework are deemed preliminary versions and are open to refinement through fieldwork and sample collection. This refinement process can produce an interpretable map aiding decisions on whether to proceed with regional exploration operations and where to collect additional ground truth data.

CNNs exhibit translation invariance, enabling them to recognise the same feature irrespective of its location in the input \citep{biscione2020learning}. This property proves especially advantageous in geological applications, where similar features may appear in different locations. The inclusion of pooling layers in CNNs facilitates dimensionality reduction, shrinking the spatial size of the input and enhancing overall network efficiency. Despite these strengths, CNNs encounter limitations in mapping spatial features. Notably, their restricted field of view implies potential oversight of relevant spatial features in the input \citep{yang2021context}, a challenge in alteration mapping with intricate and large-scale spatial patterns. Another constraint is their struggle to handle occlusions, instances where one object is partially or completely obscured by another in the input \citep{osherov2017increasing}. In the context of geological mapping via remote sensing, this might occur when one geological unit is concealed by another. Although CNNs excel at recognising spatial patterns in training data, they may grapple with generalisation to new, unseen data \citep{ren2019evaluating}, posing challenges for geological features with variations in spatial patterns across diverse regions and face challenges with different types of datasets, i.e. CNNs for ASTER data in our study.

Machine learning techniques have found widespread application in analysing remote sensing data, yet accurately predicting uncertainty remains a challenge \citep{shirmard2022review}. Estimating uncertainty associated with remote sensing models is generally difficult. The conventional approach involves validating models by comparing them with ground truth or alternative information deemed representative of ground truth. Various methods have been developed to tackle these validation challenges, resulting in the proposal of different models. Bayesian inference offers a systematic means of calculating the uncertainty of model parameters \citep{farahbakhsh2020three}. Recently, Bayesian deep learning has made significant strides, particularly in applications related to image and vision data \citep{wang2020survey,kendall2017uncertainties}. However, these methods have seen limited use in remote sensing, with their application in geological exploration being notably absent mainly due to being computationally demanding. In future research, applying Bayesian deep learning techniques \citep{chandra2022revisiting} can enhance our framework for quantifying uncertainty in geological feature predictions. Bayesian methods can address challenges such as noisy data and incomplete datasets and provide a mechanism to impute missing data \cite{chen2024deep}. Bayesian neural networks can enable uncertainty quantification from model parameters and data \cite{chandra2024bayesian}, thereby playing a pivotal role in advancing geological remote sensing and bolstering its effectiveness in mineral exploration. Moreover, incorporating ensemble machine learning and innovative data augmentation methods can address the class imbalance issue in remote sensing applications \citep{khan2023review}.

\section{Conclusions}

In this study, we introduced a framework to evaluate the effectiveness of CNNs alongside three machine learning models—KNN, SVM, and MLP—for mapping alteration zones using three distinct remote sensing datasets. Our results show that the CNN model slightly outperforms the others in mapping alteration zones across diverse datasets, especially when supported by a robust set of ground truth-based training samples. This highlights the potential of CNNs in remote sensing data processing for geological studies, particularly in critical mineral exploration, offering enhanced accuracy and efficiency. Additionally, our findings suggest that training with ground truth-based datasets yields more reliable alteration maps compared to those prepared using PCA. We also found that Landsat 9 outperforms Landsat 8 in mapping iron oxide areas when CNNs are trained with ground truth data. Moreover, using ASTER data with CNNs trained on ground truth-based datasets produced the most accurate maps for identifying argillic and propylitic alteration zones.

Despite these promising results, a significant challenge remains in validating remote sensing outputs through deep learning, which requires comprehensive ground truth datasets for reliable verification. Future research could explore crowd-sourcing as a potential solution to this challenge, leveraging its benefits for broader applications in automated geological and environmental mapping. Our alteration mapping framework has demonstrated its ability to produce accurate alteration maps and holds the potential to replace traditional methods such as spectral indices or linear classifiers. We anticipate further improvements to this framework with the continued release of data and open-source software.

\section*{Computer Code and Data Availability}

The datasets used in this study to evaluate our framework are available on Zenodo \url{https://doi.org/10.5281/zenodo.11378720}. In addition to proposing a framework, an open-source software tool is provided to implement the framework. This tool can be downloaded from \url{https://github.com/sydney-machine-learning/deeplearning_alteration_zones}.

\bibliographystyle{elsarticle-harv}
\bibliography{References}

\begin{thebibliography}{89}
\expandafter\ifx\csname natexlab\endcsname\relax\def\natexlab#1{#1}\fi
\providecommand{\url}[1]{\texttt{#1}}
\providecommand{\href}[2]{#2}
\providecommand{\path}[1]{#1}
\providecommand{\DOIprefix}{doi:}
\providecommand{\ArXivprefix}{arXiv:}
\providecommand{\URLprefix}{URL: }
\providecommand{\Pubmedprefix}{pmid:}
\providecommand{\doi}[1]{\href{http://dx.doi.org/#1}{\path{#1}}}
\providecommand{\Pubmed}[1]{\href{pmid:#1}{\path{#1}}}
\providecommand{\bibinfo}[2]{#2}
\ifx\xfnm\relax \def\xfnm[#1]{\unskip,\space#1}\fi
\bibitem[{Aamir et~al.(2019)Aamir, Rahman, Abro, Tahir and
  Ahmed}]{aamir2019optimized}
\bibinfo{author}{Aamir, M.}, \bibinfo{author}{Rahman, Z.},
  \bibinfo{author}{Abro, W.A.}, \bibinfo{author}{Tahir, M.},
  \bibinfo{author}{Ahmed, S.M.}, \bibinfo{year}{2019}.
\newblock \bibinfo{title}{An optimized architecture of image classification
  using convolutional neural network}.
\newblock \bibinfo{journal}{International Journal of Image, Graphics and Signal
  Processing} \bibinfo{volume}{10}, \bibinfo{pages}{30}.
\bibitem[{Abrams(2000)}]{abrams2000advanced}
\bibinfo{author}{Abrams, M.}, \bibinfo{year}{2000}.
\newblock \bibinfo{title}{The advanced spaceborne thermal emission and
  reflection radiometer (aster): Data products for the high spatial resolution
  imager on nasa's terra platform}.
\newblock \bibinfo{journal}{International Journal of Remote sensing}
  \bibinfo{volume}{21}, \bibinfo{pages}{847--859}.
\bibitem[{Alzubaidi et~al.(2021)Alzubaidi, Zhang, Humaidi, Al-Dujaili, Duan,
  Al-Shamma, Santamar{\'\i}a, Fadhel, Al-Amidie and
  Farhan}]{alzubaidi2021review}
\bibinfo{author}{Alzubaidi, L.}, \bibinfo{author}{Zhang, J.},
  \bibinfo{author}{Humaidi, A.J.}, \bibinfo{author}{Al-Dujaili, A.},
  \bibinfo{author}{Duan, Y.}, \bibinfo{author}{Al-Shamma, O.},
  \bibinfo{author}{Santamar{\'\i}a, J.}, \bibinfo{author}{Fadhel, M.A.},
  \bibinfo{author}{Al-Amidie, M.}, \bibinfo{author}{Farhan, L.},
  \bibinfo{year}{2021}.
\newblock \bibinfo{title}{Review of deep learning: Concepts, {CNN}
  architectures, challenges, applications, future directions}.
\newblock \bibinfo{journal}{Journal of big Data} \bibinfo{volume}{8},
  \bibinfo{pages}{1--74}.
\newblock \DOIprefix\doi{10.1186/s40537-021-00444-8}.
\bibitem[{Arulampalam and Bouzerdoum(2003)}]{arulampalam2003generalized}
\bibinfo{author}{Arulampalam, G.}, \bibinfo{author}{Bouzerdoum, A.},
  \bibinfo{year}{2003}.
\newblock \bibinfo{title}{A generalized feedforward neural network architecture
  for classification and regression}.
\newblock \bibinfo{journal}{Neural networks} \bibinfo{volume}{16},
  \bibinfo{pages}{561--568}.
\bibitem[{Bernstein et~al.(2012)Bernstein, Jin, Gregor and
  Adler-Golden}]{bernstein2012quick}
\bibinfo{author}{Bernstein, L.S.}, \bibinfo{author}{Jin, X.},
  \bibinfo{author}{Gregor, B.}, \bibinfo{author}{Adler-Golden, S.M.},
  \bibinfo{year}{2012}.
\newblock \bibinfo{title}{Quick atmospheric correction code: Algorithm
  description and recent upgrades}.
\newblock \bibinfo{journal}{Optical engineering} \bibinfo{volume}{51},
  \bibinfo{pages}{111719--111719}.
\bibitem[{Bhadra et~al.(2013)Bhadra, Pathak, Karunakar and
  Sharma}]{bhadra2013aster}
\bibinfo{author}{Bhadra, B.}, \bibinfo{author}{Pathak, S.},
  \bibinfo{author}{Karunakar, G.}, \bibinfo{author}{Sharma, J.},
  \bibinfo{year}{2013}.
\newblock \bibinfo{title}{Aster data analysis for mineral potential mapping
  around sawar-malpura area, central rajasthan}.
\newblock \bibinfo{journal}{Journal of the Indian Society of Remote Sensing}
  \bibinfo{volume}{41}, \bibinfo{pages}{391--404}.
\bibitem[{Biscione and Bowers(2020)}]{biscione2020learning}
\bibinfo{author}{Biscione, V.}, \bibinfo{author}{Bowers, J.},
  \bibinfo{year}{2020}.
\newblock \bibinfo{title}{Learning translation invariance in cnns}.
\newblock \bibinfo{journal}{arXiv} \bibinfo{volume}{arXiv:2011.11757}.
\bibitem[{Bishop and Nasrabadi(2006)}]{bishop2006pattern}
\bibinfo{author}{Bishop, C.M.}, \bibinfo{author}{Nasrabadi, N.M.},
  \bibinfo{year}{2006}.
\newblock \bibinfo{title}{Pattern Recognition and Machine Learning}.
  volume~\bibinfo{volume}{4}.
\newblock \bibinfo{publisher}{Springer}.
\bibitem[{Bradley(1997)}]{bradley1997use}
\bibinfo{author}{Bradley, A.P.}, \bibinfo{year}{1997}.
\newblock \bibinfo{title}{The use of the area under the {ROC} curve in the
  evaluation of machine learning algorithms}.
\newblock \bibinfo{journal}{Pattern recognition} \bibinfo{volume}{30},
  \bibinfo{pages}{1145--1159}.
\bibitem[{Campbell et~al.(2005)Campbell, Duncan and
  Hibbits}]{campbell2005analysis}
\bibinfo{author}{Campbell, E.}, \bibinfo{author}{Duncan, I.},
  \bibinfo{author}{Hibbits, H.}, \bibinfo{year}{2005}.
\newblock \bibinfo{title}{Analysis of errors occurring in the transfer of
  geologic point data from field maps to digital data sets}, in:
  \bibinfo{booktitle}{USGS Workshop on Digital Mapping Techniques},
  p.~\bibinfo{pages}{61}.
\bibitem[{Chandra et~al.(2022)Chandra, Jain, Maharana and
  Krivitsky}]{chandra2022revisiting}
\bibinfo{author}{Chandra, R.}, \bibinfo{author}{Jain, M.},
  \bibinfo{author}{Maharana, M.}, \bibinfo{author}{Krivitsky, P.N.},
  \bibinfo{year}{2022}.
\newblock \bibinfo{title}{Revisiting bayesian autoencoders with mcmc}.
\newblock \bibinfo{journal}{IEEE Access} \bibinfo{volume}{10},
  \bibinfo{pages}{40482--40495}.
\bibitem[{Chandra and Simmons(2024)}]{chandra2024bayesian}
\bibinfo{author}{Chandra, R.}, \bibinfo{author}{Simmons, J.},
  \bibinfo{year}{2024}.
\newblock \bibinfo{title}{Bayesian neural networks via mcmc: A python-based
  tutorial}.
\newblock \bibinfo{journal}{IEEE Access} \bibinfo{volume}{12},
  \bibinfo{pages}{70519--70549}.
\newblock \DOIprefix\doi{10.1109/ACCESS.2024.3401234}.
\bibitem[{Chantry et~al.(2021)Chantry, Christensen, Dueben and
  Palmer}]{chantry2021opportunities}
\bibinfo{author}{Chantry, M.}, \bibinfo{author}{Christensen, H.},
  \bibinfo{author}{Dueben, P.}, \bibinfo{author}{Palmer, T.},
  \bibinfo{year}{2021}.
\newblock \bibinfo{title}{Opportunities and challenges for machine learning in
  weather and climate modelling: hard, medium and soft ai}.
\newblock \bibinfo{journal}{Philosophical Transactions of the Royal Society A:
  Mathematical, Physical and Engineering Sciences} \bibinfo{volume}{379},
  \bibinfo{pages}{20200083}.
\bibitem[{Chen et~al.(2024)Chen, Andersen and Chandra}]{chen2024deep}
\bibinfo{author}{Chen, E.}, \bibinfo{author}{Andersen, M.S.},
  \bibinfo{author}{Chandra, R.}, \bibinfo{year}{2024}.
\newblock \bibinfo{title}{Deep learning framework with bayesian data imputation
  for modelling and forecasting groundwater levels}.
\newblock \bibinfo{journal}{Environmental Modelling \& Software} ,
  \bibinfo{pages}{106072}\DOIprefix\doi{10.1016/j.envsoft.2024.106072}.
\bibitem[{Chen et~al.(2018)Chen, Cai and Li}]{chen2018recognition}
\bibinfo{author}{Chen, G.}, \bibinfo{author}{Cai, Z.}, \bibinfo{author}{Li,
  X.}, \bibinfo{year}{2018}.
\newblock \bibinfo{title}{Recognition and classification of high resolution
  remote sensing image based on convolutional neural network}.
\newblock \bibinfo{journal}{International Journal of Performability
  Engineering} \bibinfo{volume}{14}, \bibinfo{pages}{2852}.
\bibitem[{Chirico et~al.(2022)Chirico, Mondillo, Laukamp, Mormone, Di~Martire,
  Novellino and Balassone}]{chirico2022mapping}
\bibinfo{author}{Chirico, R.}, \bibinfo{author}{Mondillo, N.},
  \bibinfo{author}{Laukamp, C.}, \bibinfo{author}{Mormone, A.},
  \bibinfo{author}{Di~Martire, D.}, \bibinfo{author}{Novellino, A.},
  \bibinfo{author}{Balassone, G.}, \bibinfo{year}{2022}.
\newblock \bibinfo{title}{Mapping hydrothermal and supergene alteration zones
  associated with carbonate-hosted zn-pb deposits by using prisma satellite
  imagery supported by field-based hyperspectral data, mineralogical and
  geochemical analysis}.
\newblock \bibinfo{journal}{Ore Geology Reviews} , \bibinfo{pages}{105244}.
\bibitem[{Farahbakhsh et~al.(2020)Farahbakhsh, Hezarkhani, Eslamkish, Bahroudi
  and Chandra}]{farahbakhsh2020three}
\bibinfo{author}{Farahbakhsh, E.}, \bibinfo{author}{Hezarkhani, A.},
  \bibinfo{author}{Eslamkish, T.}, \bibinfo{author}{Bahroudi, A.},
  \bibinfo{author}{Chandra, R.}, \bibinfo{year}{2020}.
\newblock \bibinfo{title}{Three-dimensional weights of evidence modelling of a
  deep-seated porphyry cu deposit}.
\newblock \bibinfo{journal}{Geochemistry: Exploration, Environment, Analysis}
  \bibinfo{volume}{20}, \bibinfo{pages}{480--495}.
\bibitem[{Farahbakhsh et~al.(2016)Farahbakhsh, Shirmard, Bahroudi and
  Eslamkish}]{farahbakhsh2016fusing}
\bibinfo{author}{Farahbakhsh, E.}, \bibinfo{author}{Shirmard, H.},
  \bibinfo{author}{Bahroudi, A.}, \bibinfo{author}{Eslamkish, T.},
  \bibinfo{year}{2016}.
\newblock \bibinfo{title}{Fusing aster and quickbird-2 satellite data for
  detailed investigation of porphyry copper deposits using pca; case study of
  naysian deposit, iran}.
\newblock \bibinfo{journal}{Journal of the Indian Society of Remote Sensing}
  \bibinfo{volume}{44}, \bibinfo{pages}{525--537}.
\bibitem[{Fergusson and Henderson(2015)}]{fergusson2015early}
\bibinfo{author}{Fergusson, C.L.}, \bibinfo{author}{Henderson, R.},
  \bibinfo{year}{2015}.
\newblock \bibinfo{title}{Early palaeozoic continental growth in the tasmanides
  of northeast gondwana and its implications for rodinia assembly and rifting}.
\newblock \bibinfo{journal}{Gondwana Research} \bibinfo{volume}{28},
  \bibinfo{pages}{933--953}.
\bibitem[{He et~al.(2010)He, He and Cui}]{he2010hydrothermal}
\bibinfo{author}{He, Z.}, \bibinfo{author}{He, B.}, \bibinfo{author}{Cui, Y.},
  \bibinfo{year}{2010}.
\newblock \bibinfo{title}{Hydrothermal alteration mapping using aster data in
  east kunlun mountains, china}, in: \bibinfo{booktitle}{IEEE International
  Geoscience and Remote Sensing Symposium}, pp. \bibinfo{pages}{4514--4517}.
\bibitem[{Ienco et~al.(2017)Ienco, Gaetano, Dupaquier and
  Maurel}]{ienco2017land}
\bibinfo{author}{Ienco, D.}, \bibinfo{author}{Gaetano, R.},
  \bibinfo{author}{Dupaquier, C.}, \bibinfo{author}{Maurel, P.},
  \bibinfo{year}{2017}.
\newblock \bibinfo{title}{Land cover classification via multitemporal spatial
  data by deep recurrent neural networks}.
\newblock \bibinfo{journal}{IEEE Geoscience and Remote Sensing Letters}
  \bibinfo{volume}{14}, \bibinfo{pages}{1685--1689}.
\bibitem[{Ireland et~al.(1998)Ireland, Flottmann, Fanning, Gibson and
  Preiss}]{ireland1998development}
\bibinfo{author}{Ireland, T.}, \bibinfo{author}{Flottmann, T.},
  \bibinfo{author}{Fanning, C.}, \bibinfo{author}{Gibson, G.},
  \bibinfo{author}{Preiss, W.V.}, \bibinfo{year}{1998}.
\newblock \bibinfo{title}{Development of the early paleozoic pacific margin of
  gondwana from detrital-zircon ages across the delamerian orogen}.
\newblock \bibinfo{journal}{Geology} \bibinfo{volume}{26},
  \bibinfo{pages}{243--246}.
\bibitem[{Jolliffe and Cadima(2016)}]{jolliffe2016principal}
\bibinfo{author}{Jolliffe, I.T.}, \bibinfo{author}{Cadima, J.},
  \bibinfo{year}{2016}.
\newblock \bibinfo{title}{Principal component analysis: a review and recent
  developments}.
\newblock \bibinfo{journal}{Philosophical transactions of the royal society A:
  Mathematical, Physical and Engineering Sciences} \bibinfo{volume}{374},
  \bibinfo{pages}{20150202}.
\bibitem[{Karpathy and Fei-Fei(2015)}]{karpathy2015deep}
\bibinfo{author}{Karpathy, A.}, \bibinfo{author}{Fei-Fei, L.},
  \bibinfo{year}{2015}.
\newblock \bibinfo{title}{Deep visual-semantic alignments for generating image
  descriptions}, in: \bibinfo{booktitle}{IEEE Conference on Computer Vision and
  Pattern Recognition}, pp. \bibinfo{pages}{3128--3137}.
\bibitem[{Kattenborn et~al.(2021)Kattenborn, Leitloff, Schiefer and
  Hinz}]{kattenborn2021review}
\bibinfo{author}{Kattenborn, T.}, \bibinfo{author}{Leitloff, J.},
  \bibinfo{author}{Schiefer, F.}, \bibinfo{author}{Hinz, S.},
  \bibinfo{year}{2021}.
\newblock \bibinfo{title}{Review on convolutional neural networks (cnn) in
  vegetation remote sensing}.
\newblock \bibinfo{journal}{ISPRS journal of photogrammetry and remote sensing}
  \bibinfo{volume}{173}, \bibinfo{pages}{24--49}.
\newblock \DOIprefix\doi{10.1016/j.isprsjprs.2020.12.010}.
\bibitem[{Kavitha et~al.(2014)Kavitha, Nachammai, Ranjani and
  Shifali}]{kavitha2014speech}
\bibinfo{author}{Kavitha, R.}, \bibinfo{author}{Nachammai, N.},
  \bibinfo{author}{Ranjani, R.}, \bibinfo{author}{Shifali, J.},
  \bibinfo{year}{2014}.
\newblock \bibinfo{title}{Speech based voice recognition system for natural
  language processing}.
\newblock \bibinfo{journal}{International Journal of Computer Science and
  Information Technologies} \bibinfo{volume}{5}, \bibinfo{pages}{5301--5305}.
\bibitem[{Kendall and Gal(2017)}]{kendall2017uncertainties}
\bibinfo{author}{Kendall, A.}, \bibinfo{author}{Gal, Y.}, \bibinfo{year}{2017}.
\newblock \bibinfo{title}{What uncertainties do we need in bayesian deep
  learning for computer vision?}
\newblock \bibinfo{journal}{Advances in neural information processing systems}
  \bibinfo{volume}{30}.
\bibitem[{Kesler(2007)}]{kesler2007mineral}
\bibinfo{author}{Kesler, S.E.}, \bibinfo{year}{2007}.
\newblock \bibinfo{title}{Mineral supply and demand into the 21st century}, in:
  \bibinfo{booktitle}{USGS workshop on Deposit Modeling, Mineral Resource
  Assessment, and Their Role in Sustainable Development}, pp.
  \bibinfo{pages}{55--62}.
\bibitem[{Khan et~al.(2023)Khan, Chaudhari and Chandra}]{khan2023review}
\bibinfo{author}{Khan, A.A.}, \bibinfo{author}{Chaudhari, O.},
  \bibinfo{author}{Chandra, R.}, \bibinfo{year}{2023}.
\newblock \bibinfo{title}{A review of ensemble learning and data augmentation
  models for class imbalanced problems: combination, implementation and
  evaluation}.
\newblock \bibinfo{journal}{arXiv} \bibinfo{volume}{arXiv:2304.02858}.
\bibitem[{Kingma and Ba(2014)}]{kingma2014adam}
\bibinfo{author}{Kingma, D.P.}, \bibinfo{author}{Ba, J.}, \bibinfo{year}{2014}.
\newblock \bibinfo{title}{Adam: A method for stochastic optimization}.
\newblock \bibinfo{journal}{arXiv} \bibinfo{volume}{arXiv:1412.6980}.
\bibitem[{Klambauer et~al.(2017)Klambauer, Unterthiner, Mayr and
  Hochreiter}]{klambauer2017self}
\bibinfo{author}{Klambauer, G.}, \bibinfo{author}{Unterthiner, T.},
  \bibinfo{author}{Mayr, A.}, \bibinfo{author}{Hochreiter, S.},
  \bibinfo{year}{2017}.
\newblock \bibinfo{title}{Self-normalizing neural networks}.
\newblock \bibinfo{journal}{Advances in Neural Information Processing Systems}
  \bibinfo{volume}{30}.
\bibitem[{Kotsiantis et~al.(2006)Kotsiantis, Zaharakis and
  Pintelas}]{kotsiantis2006machine}
\bibinfo{author}{Kotsiantis, S.B.}, \bibinfo{author}{Zaharakis, I.D.},
  \bibinfo{author}{Pintelas, P.E.}, \bibinfo{year}{2006}.
\newblock \bibinfo{title}{Machine learning: a review of classification and
  combining techniques}.
\newblock \bibinfo{journal}{Artificial Intelligence Review}
  \bibinfo{volume}{26}, \bibinfo{pages}{159--190}.
\bibitem[{Kreinovich and Kosheleva(2020)}]{kreinovich2020deep}
\bibinfo{author}{Kreinovich, V.}, \bibinfo{author}{Kosheleva, O.},
  \bibinfo{year}{2020}.
\newblock \bibinfo{title}{Deep learning (partly) demystified}, in:
  \bibinfo{booktitle}{4th International Conference on Intelligent Systems,
  Metaheuristics \& Swarm Intelligence}, pp. \bibinfo{pages}{30--35}.
\bibitem[{{L3Harris Geospatial}(2022)}]{harris2022envi}
\bibinfo{author}{{L3Harris Geospatial}}, \bibinfo{year}{2022}.
\newblock \bibinfo{title}{Envi 5.6}.
\newblock \URLprefix
  \url{https://www.l3harrisgeospatial.com/Software-Technology/ENVI}.
\bibitem[{Large et~al.(2005)Large, Bull, McGoldrick, Walters, Derrick and
  Carr}]{large2005stratiform}
\bibinfo{author}{Large, R.R.}, \bibinfo{author}{Bull, S.W.},
  \bibinfo{author}{McGoldrick, P.J.}, \bibinfo{author}{Walters, S.},
  \bibinfo{author}{Derrick, G.M.}, \bibinfo{author}{Carr, G.R.},
  \bibinfo{year}{2005}.
\newblock \bibinfo{title}{Stratiform and strata-bound zn-pb-ag deposits in
  proterozoic sedimentary basins, northern australia}, in:
  \bibinfo{booktitle}{One Hundredth Anniversary Volume}.
  \bibinfo{publisher}{Society of Economic Geologists}.
\bibitem[{Lee et~al.(2014)Lee, Sun and Saunders}]{lee2014proximal}
\bibinfo{author}{Lee, J.D.}, \bibinfo{author}{Sun, Y.},
  \bibinfo{author}{Saunders, M.A.}, \bibinfo{year}{2014}.
\newblock \bibinfo{title}{Proximal newton-type methods for minimizing composite
  functions}.
\newblock \bibinfo{journal}{SIAM Journal on Optimization} \bibinfo{volume}{24},
  \bibinfo{pages}{1420--1443}.
\bibitem[{Li et~al.(2019)Li, Wang, Zhang, Gao, Yang and Wang}]{li2019deep}
\bibinfo{author}{Li, C.}, \bibinfo{author}{Wang, Y.}, \bibinfo{author}{Zhang,
  X.}, \bibinfo{author}{Gao, H.}, \bibinfo{author}{Yang, Y.},
  \bibinfo{author}{Wang, J.}, \bibinfo{year}{2019}.
\newblock \bibinfo{title}{Deep belief network for spectral--spatial
  classification of hyperspectral remote sensor data}.
\newblock \bibinfo{journal}{Sensors} \bibinfo{volume}{19},
  \bibinfo{pages}{204}.
\bibitem[{Li et~al.(2022a)Li, Wu, Wang, Lan and Xiao}]{li2022stm}
\bibinfo{author}{Li, P.}, \bibinfo{author}{Wu, J.}, \bibinfo{author}{Wang, Y.},
  \bibinfo{author}{Lan, Q.}, \bibinfo{author}{Xiao, W.}, \bibinfo{year}{2022}a.
\newblock \bibinfo{title}{Stm: Spectrogram transformer model for underwater
  acoustic target recognition}.
\newblock \bibinfo{journal}{Journal of Marine Science and Engineering}
  \bibinfo{volume}{10}, \bibinfo{pages}{1428}.
\bibitem[{Li et~al.(2022b)Li, Liu, Yang, Peng and Zhou}]{li2021survey}
\bibinfo{author}{Li, Z.}, \bibinfo{author}{Liu, F.}, \bibinfo{author}{Yang,
  W.}, \bibinfo{author}{Peng, S.}, \bibinfo{author}{Zhou, J.},
  \bibinfo{year}{2022}b.
\newblock \bibinfo{title}{A survey of convolutional neural networks: Analysis,
  applications, and prospects}.
\newblock \bibinfo{journal}{IEEE Transactions on Neural Networks and Learning
  Systems} \bibinfo{volume}{33}, \bibinfo{pages}{6999--7019}.
\bibitem[{Linardos et~al.(2022)Linardos, Drakaki, Tzionas and
  Karnavas}]{linardos2022machine}
\bibinfo{author}{Linardos, V.}, \bibinfo{author}{Drakaki, M.},
  \bibinfo{author}{Tzionas, P.}, \bibinfo{author}{Karnavas, Y.L.},
  \bibinfo{year}{2022}.
\newblock \bibinfo{title}{Machine learning in disaster management: Recent
  developments in methods and applications}.
\newblock \bibinfo{journal}{Machine Learning and Knowledge Extraction}
  \bibinfo{volume}{4}, \bibinfo{pages}{446--473}.
\bibitem[{Liu et~al.(2022)Liu, Tang and Huang}]{liu2022building}
\bibinfo{author}{Liu, Z.}, \bibinfo{author}{Tang, H.}, \bibinfo{author}{Huang,
  W.}, \bibinfo{year}{2022}.
\newblock \bibinfo{title}{Building outline delineation from vhr remote sensing
  images using the convolutional recurrent neural network embedded with line
  segment information}.
\newblock \bibinfo{journal}{IEEE Transactions on Geoscience and Remote Sensing}
  \bibinfo{volume}{60}, \bibinfo{pages}{1--13}.
\bibitem[{Lyons et~al.(2018)Lyons, Keith, Phinn, Mason and
  Elith}]{lyons2018comparison}
\bibinfo{author}{Lyons, M.B.}, \bibinfo{author}{Keith, D.A.},
  \bibinfo{author}{Phinn, S.R.}, \bibinfo{author}{Mason, T.J.},
  \bibinfo{author}{Elith, J.}, \bibinfo{year}{2018}.
\newblock \bibinfo{title}{A comparison of resampling methods for remote sensing
  classification and accuracy assessment}.
\newblock \bibinfo{journal}{Remote Sensing of Environment}
  \bibinfo{volume}{208}, \bibinfo{pages}{145--153}.
\bibitem[{Ma et~al.(2019a)Ma, Sun, Deng, Huang, Tao, Zhu, Teng and
  Meng}]{ma2019evaluation}
\bibinfo{author}{Ma, J.}, \bibinfo{author}{Sun, Y.}, \bibinfo{author}{Deng,
  G.}, \bibinfo{author}{Huang, S.}, \bibinfo{author}{Tao, Y.},
  \bibinfo{author}{Zhu, H.}, \bibinfo{author}{Teng, Q.}, \bibinfo{author}{Meng,
  X.}, \bibinfo{year}{2019}a.
\newblock \bibinfo{title}{Evaluation of different approaches of convolutional
  neural networks for land use and land cover classification based on high
  resolution remote sensing images}, in: \bibinfo{booktitle}{IEEE International
  Conference on Signal, Information and Data Processing}, pp.
  \bibinfo{pages}{1--4}.
\bibitem[{Ma et~al.(2019b)Ma, Liu, Zhang, Ye, Yin and Johnson}]{ma2019deep}
\bibinfo{author}{Ma, L.}, \bibinfo{author}{Liu, Y.}, \bibinfo{author}{Zhang,
  X.}, \bibinfo{author}{Ye, Y.}, \bibinfo{author}{Yin, G.},
  \bibinfo{author}{Johnson, B.A.}, \bibinfo{year}{2019}b.
\newblock \bibinfo{title}{Deep learning in remote sensing applications: A
  meta-analysis and review}.
\newblock \bibinfo{journal}{ISPRS journal of photogrammetry and remote sensing}
  \bibinfo{volume}{152}, \bibinfo{pages}{166--177}.
\bibitem[{Maggiori et~al.(2016)Maggiori, Tarabalka, Charpiat and
  Alliez}]{maggiori2016convolutional}
\bibinfo{author}{Maggiori, E.}, \bibinfo{author}{Tarabalka, Y.},
  \bibinfo{author}{Charpiat, G.}, \bibinfo{author}{Alliez, P.},
  \bibinfo{year}{2016}.
\newblock \bibinfo{title}{Convolutional neural networks for large-scale
  remote-sensing image classification}.
\newblock \bibinfo{journal}{IEEE Transactions on geoscience and remote sensing}
  \bibinfo{volume}{55}, \bibinfo{pages}{645--657}.
\bibitem[{Maggiori et~al.(2017)Maggiori, Tarabalka, Charpiat and
  Alliez}]{maggiori2017convolutional}
\bibinfo{author}{Maggiori, E.}, \bibinfo{author}{Tarabalka, Y.},
  \bibinfo{author}{Charpiat, G.}, \bibinfo{author}{Alliez, P.},
  \bibinfo{year}{2017}.
\newblock \bibinfo{title}{Convolutional neural networks for large-scale
  remote-sensing image classification}.
\newblock \bibinfo{journal}{IEEE Transactions on Geoscience and Remote Sensing}
  \bibinfo{volume}{55}, \bibinfo{pages}{645--657}.
\newblock \DOIprefix\doi{10.1109/TGRS.2016.2612821}.
\bibitem[{Maleki et~al.(2021)Maleki, Niroomand, Farahbakhsh, Modabberi and
  Tajeddin}]{maleki2021hydrothermal}
\bibinfo{author}{Maleki, M.}, \bibinfo{author}{Niroomand, S.},
  \bibinfo{author}{Farahbakhsh, E.}, \bibinfo{author}{Modabberi, S.},
  \bibinfo{author}{Tajeddin, H.A.}, \bibinfo{year}{2021}.
\newblock \bibinfo{title}{Hydrothermal alteration and structural mapping of the
  qolqoleh-kasnazan shear zone in iran using remote sensing data}.
\newblock \bibinfo{journal}{Arabian Journal of Geosciences}
  \bibinfo{volume}{14}, \bibinfo{pages}{1--14}.
\bibitem[{McCuaig et~al.(2010)McCuaig, Beresford and
  Hronsky}]{mccuaig2010translating}
\bibinfo{author}{McCuaig, T.C.}, \bibinfo{author}{Beresford, S.},
  \bibinfo{author}{Hronsky, J.}, \bibinfo{year}{2010}.
\newblock \bibinfo{title}{Translating the mineral systems approach into an
  effective exploration targeting system}.
\newblock \bibinfo{journal}{Ore Geology Reviews} \bibinfo{volume}{38},
  \bibinfo{pages}{128--138}.
\bibitem[{Van~der Meer et~al.(2012)Van~der Meer, Van~der Werff, Van~Ruitenbeek,
  Hecker, Bakker, Noomen, Van Der~Meijde, Carranza, De~Smeth and
  Woldai}]{van2012multi}
\bibinfo{author}{Van~der Meer, F.D.}, \bibinfo{author}{Van~der Werff, H.M.},
  \bibinfo{author}{Van~Ruitenbeek, F.J.}, \bibinfo{author}{Hecker, C.A.},
  \bibinfo{author}{Bakker, W.H.}, \bibinfo{author}{Noomen, M.F.},
  \bibinfo{author}{Van Der~Meijde, M.}, \bibinfo{author}{Carranza, E.J.M.},
  \bibinfo{author}{De~Smeth, J.B.}, \bibinfo{author}{Woldai, T.},
  \bibinfo{year}{2012}.
\newblock \bibinfo{title}{Multi-and hyperspectral geologic remote sensing: A
  review}.
\newblock \bibinfo{journal}{International Journal of Applied Earth Observation
  and Geoinformation} \bibinfo{volume}{14}, \bibinfo{pages}{112--128}.
\bibitem[{Mohamed et~al.(2021)Mohamed, Al-Naimi, Mgbeojedo and
  Agoha}]{mohamed2021geological}
\bibinfo{author}{Mohamed, M.T.A.}, \bibinfo{author}{Al-Naimi, L.S.},
  \bibinfo{author}{Mgbeojedo, T.I.}, \bibinfo{author}{Agoha, C.C.},
  \bibinfo{year}{2021}.
\newblock \bibinfo{title}{Geological mapping and mineral prospectivity using
  remote sensing and gis in parts of hamissana, northeast sudan}.
\newblock \bibinfo{journal}{Journal of Petroleum Exploration and Production}
  \bibinfo{volume}{11}, \bibinfo{pages}{1123--1138}.
\bibitem[{Morland and Webster(1998)}]{morland1998broken}
\bibinfo{author}{Morland, R.}, \bibinfo{author}{Webster, A.},
  \bibinfo{year}{1998}.
\newblock \bibinfo{title}{Broken hill lead-zinc-silver deposit}.
\newblock \bibinfo{journal}{Geology of Australian and Papua New Guinean Mineral
  Deposits} , \bibinfo{pages}{619--626}.
\bibitem[{Nagi et~al.(2011)Nagi, Ducatelle, Di~Caro, Cireşan, Meier, Giusti,
  Nagi, Schmidhuber and Gambardella}]{nagi2011maxpooling}
\bibinfo{author}{Nagi, J.}, \bibinfo{author}{Ducatelle, F.},
  \bibinfo{author}{Di~Caro, G.A.}, \bibinfo{author}{Cireşan, D.},
  \bibinfo{author}{Meier, U.}, \bibinfo{author}{Giusti, A.},
  \bibinfo{author}{Nagi, F.}, \bibinfo{author}{Schmidhuber, J.},
  \bibinfo{author}{Gambardella, L.M.}, \bibinfo{year}{2011}.
\newblock \bibinfo{title}{Max-pooling convolutional neural networks for
  vision-based hand gesture recognition}, in: \bibinfo{booktitle}{IEEE
  International Conference on Signal and Image Processing Applications}, pp.
  \bibinfo{pages}{342--347}.
\newblock \DOIprefix\doi{10.1109/ICSIPA.2011.6144164}.
\bibitem[{Nair and Hinton(2010)}]{nair2010rectified}
\bibinfo{author}{Nair, V.}, \bibinfo{author}{Hinton, G.E.},
  \bibinfo{year}{2010}.
\newblock \bibinfo{title}{Rectified linear units improve restricted boltzmann
  machines}, in: \bibinfo{booktitle}{27th International Conference on Machine
  Learning}, pp. \bibinfo{pages}{807--814}.
\bibitem[{Niroumand-Jadidi et~al.(2022)Niroumand-Jadidi, Legleiter and
  Bovolo}]{niroumand2022river}
\bibinfo{author}{Niroumand-Jadidi, M.}, \bibinfo{author}{Legleiter, C.J.},
  \bibinfo{author}{Bovolo, F.}, \bibinfo{year}{2022}.
\newblock \bibinfo{title}{River bathymetry retrieval from landsat-9 images
  based on neural networks and comparison to superdove and sentinel-2}.
\newblock \bibinfo{journal}{IEEE Journal of Selected Topics in Applied Earth
  Observations and Remote Sensing} \bibinfo{volume}{15},
  \bibinfo{pages}{5250--5260}.
\bibitem[{Noble(2006)}]{noble2006support}
\bibinfo{author}{Noble, W.S.}, \bibinfo{year}{2006}.
\newblock \bibinfo{title}{What is a support vector machine?}
\newblock \bibinfo{journal}{Nature biotechnology} \bibinfo{volume}{24},
  \bibinfo{pages}{1565--1567}.
\bibitem[{Osherov and Lindenbaum(2017)}]{osherov2017increasing}
\bibinfo{author}{Osherov, E.}, \bibinfo{author}{Lindenbaum, M.},
  \bibinfo{year}{2017}.
\newblock \bibinfo{title}{Increasing cnn robustness to occlusions by reducing
  filter support}, in: \bibinfo{booktitle}{IEEE International Conference on
  Computer Vision}, pp. \bibinfo{pages}{550--561}.
\bibitem[{Pacey et~al.(2016)Pacey, Wilkinson, Boyce and
  Cooke}]{pacey2016propylitic}
\bibinfo{author}{Pacey, A.}, \bibinfo{author}{Wilkinson, J.J.},
  \bibinfo{author}{Boyce, A.J.}, \bibinfo{author}{Cooke, D.R.},
  \bibinfo{year}{2016}.
\newblock \bibinfo{title}{Propylitic alteration and metal mobility in porphyry
  systems: a case study of the northparkes cu-au deposits, nsw, australia}.
\newblock \bibinfo{journal}{Applied Earth Science} \bibinfo{volume}{125},
  \bibinfo{pages}{93--93}.
\bibitem[{Parr et~al.(2004)Parr, Stevens, Carr and Page}]{parr2004subseafloor}
\bibinfo{author}{Parr, J.M.}, \bibinfo{author}{Stevens, B.P.},
  \bibinfo{author}{Carr, G.R.}, \bibinfo{author}{Page, R.W.},
  \bibinfo{year}{2004}.
\newblock \bibinfo{title}{Subseafloor origin for broken hill pb-zn-ag
  mineralization, new south wales, australia}.
\newblock \bibinfo{journal}{Geology} \bibinfo{volume}{32},
  \bibinfo{pages}{589--592}.
\bibitem[{Pelletier et~al.(2019)Pelletier, Webb and
  Petitjean}]{pelletier2019deep}
\bibinfo{author}{Pelletier, C.}, \bibinfo{author}{Webb, G.I.},
  \bibinfo{author}{Petitjean, F.}, \bibinfo{year}{2019}.
\newblock \bibinfo{title}{Deep learning for the classification of sentinel-2
  image time series}, in: \bibinfo{booktitle}{IEEE International Geoscience and
  Remote Sensing Symposium}, pp. \bibinfo{pages}{461--464}.
\bibitem[{Plimer(1978)}]{plimer1978proximal}
\bibinfo{author}{Plimer, I.}, \bibinfo{year}{1978}.
\newblock \bibinfo{title}{Proximal and distal stratabound ore deposits}.
\newblock \bibinfo{journal}{Mineralium Deposita} \bibinfo{volume}{13},
  \bibinfo{pages}{345--353}.
\bibitem[{Pour and Hashim(2015)}]{pour2015hydrothermal}
\bibinfo{author}{Pour, A.B.}, \bibinfo{author}{Hashim, M.},
  \bibinfo{year}{2015}.
\newblock \bibinfo{title}{Hydrothermal alteration mapping from landsat-8 data,
  sar cheshmeh copper mining district, south-eastern islamic republic of iran}.
\newblock \bibinfo{journal}{Journal of Taibah University for Science}
  \bibinfo{volume}{9}, \bibinfo{pages}{155--166}.
\bibitem[{Rawat and Wang(2017)}]{rawat2017deep}
\bibinfo{author}{Rawat, W.}, \bibinfo{author}{Wang, Z.}, \bibinfo{year}{2017}.
\newblock \bibinfo{title}{Deep convolutional neural networks for image
  classification: A comprehensive review}.
\newblock \bibinfo{journal}{Neural computation} \bibinfo{volume}{29},
  \bibinfo{pages}{2352--2449}.
\bibitem[{Ren et~al.(2019)Ren, Su and Lu}]{ren2019evaluating}
\bibinfo{author}{Ren, H.}, \bibinfo{author}{Su, J.}, \bibinfo{author}{Lu, H.},
  \bibinfo{year}{2019}.
\newblock \bibinfo{title}{Evaluating generalization ability of convolutional
  neural networks and capsule networks for image classification via top-2
  classification}.
\newblock \bibinfo{journal}{arXiv} \bibinfo{volume}{arXiv:1901.10112}.
\bibitem[{Rowan and Mars(2003)}]{rowan2003lithologic}
\bibinfo{author}{Rowan, L.C.}, \bibinfo{author}{Mars, J.C.},
  \bibinfo{year}{2003}.
\newblock \bibinfo{title}{Lithologic mapping in the mountain pass, california
  area using advanced spaceborne thermal emission and reflection radiometer
  (aster) data}.
\newblock \bibinfo{journal}{Remote sensing of Environment}
  \bibinfo{volume}{84}, \bibinfo{pages}{350--366}.
\bibitem[{Roy et~al.(2014)Roy, Wulder, Loveland, Woodcock, Allen, Anderson,
  Helder, Irons, Johnson, Kennedy et~al.}]{roy2014landsat}
\bibinfo{author}{Roy, D.P.}, \bibinfo{author}{Wulder, M.A.},
  \bibinfo{author}{Loveland, T.R.}, \bibinfo{author}{Woodcock, C.E.},
  \bibinfo{author}{Allen, R.G.}, \bibinfo{author}{Anderson, M.C.},
  \bibinfo{author}{Helder, D.}, \bibinfo{author}{Irons, J.R.},
  \bibinfo{author}{Johnson, D.M.}, \bibinfo{author}{Kennedy, R.}, et~al.,
  \bibinfo{year}{2014}.
\newblock \bibinfo{title}{Landsat-8: Science and product vision for terrestrial
  global change research}.
\newblock \bibinfo{journal}{Remote sensing of Environment}
  \bibinfo{volume}{145}, \bibinfo{pages}{154--172}.
\bibitem[{Sabins(1999)}]{sabins1999remote}
\bibinfo{author}{Sabins, F.F.}, \bibinfo{year}{1999}.
\newblock \bibinfo{title}{Remote sensing for mineral exploration}.
\newblock \bibinfo{journal}{Ore geology reviews} \bibinfo{volume}{14},
  \bibinfo{pages}{157--183}.
\bibitem[{Scholkopf et~al.(1997)Scholkopf, Sung, Burges, Girosi, Niyogi, Poggio
  and Vapnik}]{scholkopf1997comparing}
\bibinfo{author}{Scholkopf, B.}, \bibinfo{author}{Sung, K.K.},
  \bibinfo{author}{Burges, C.}, \bibinfo{author}{Girosi, F.},
  \bibinfo{author}{Niyogi, P.}, \bibinfo{author}{Poggio, T.},
  \bibinfo{author}{Vapnik, V.}, \bibinfo{year}{1997}.
\newblock \bibinfo{title}{Comparing support vector machines with gaussian
  kernels to radial basis function classifiers}.
\newblock \bibinfo{journal}{IEEE Transactions on Signal Processing}
  \bibinfo{volume}{45}, \bibinfo{pages}{2758--2765}.
\newblock \DOIprefix\doi{10.1109/78.650102}.
\bibitem[{Shakya et~al.(2021)Shakya, Biswas and Pal}]{shakya2021parametric}
\bibinfo{author}{Shakya, A.}, \bibinfo{author}{Biswas, M.},
  \bibinfo{author}{Pal, M.}, \bibinfo{year}{2021}.
\newblock \bibinfo{title}{Parametric study of convolutional neural network
  based remote sensing image classification}.
\newblock \bibinfo{journal}{International Journal of Remote Sensing}
  \bibinfo{volume}{42}, \bibinfo{pages}{2663--2685}.
\bibitem[{Shirmard et~al.(2020)Shirmard, Farahbakhsh, Beiranvand~Pour, Muslim,
  M{\"u}ller and Chandra}]{shirmard2020integration}
\bibinfo{author}{Shirmard, H.}, \bibinfo{author}{Farahbakhsh, E.},
  \bibinfo{author}{Beiranvand~Pour, A.}, \bibinfo{author}{Muslim, A.M.},
  \bibinfo{author}{M{\"u}ller, R.D.}, \bibinfo{author}{Chandra, R.},
  \bibinfo{year}{2020}.
\newblock \bibinfo{title}{Integration of selective dimensionality reduction
  techniques for mineral exploration using aster satellite data}.
\newblock \bibinfo{journal}{Remote Sensing} \bibinfo{volume}{12},
  \bibinfo{pages}{1261}.
\bibitem[{Shirmard et~al.(2022)Shirmard, Farahbakhsh, M{\"u}ller and
  Chandra}]{shirmard2022review}
\bibinfo{author}{Shirmard, H.}, \bibinfo{author}{Farahbakhsh, E.},
  \bibinfo{author}{M{\"u}ller, R.D.}, \bibinfo{author}{Chandra, R.},
  \bibinfo{year}{2022}.
\newblock \bibinfo{title}{A review of machine learning in processing remote
  sensing data for mineral exploration}.
\newblock \bibinfo{journal}{Remote Sensing of Environment}
  \bibinfo{volume}{268}, \bibinfo{pages}{112750}.
\bibitem[{Sillitoe et~al.(1998)Sillitoe, Steele, Thompson and
  Lang}]{sillitoe1998advanced}
\bibinfo{author}{Sillitoe, R.}, \bibinfo{author}{Steele, G.},
  \bibinfo{author}{Thompson, J.}, \bibinfo{author}{Lang, J.},
  \bibinfo{year}{1998}.
\newblock \bibinfo{title}{Advanced argillic lithocaps in the bolivian
  tin-silver belt}.
\newblock \bibinfo{journal}{Mineralium Deposita} \bibinfo{volume}{33},
  \bibinfo{pages}{539--546}.
\bibitem[{Song et~al.(2019)Song, Gao, Zhu and Ma}]{song2019survey}
\bibinfo{author}{Song, J.}, \bibinfo{author}{Gao, S.}, \bibinfo{author}{Zhu,
  Y.}, \bibinfo{author}{Ma, C.}, \bibinfo{year}{2019}.
\newblock \bibinfo{title}{A survey of remote sensing image classification based
  on cnns}.
\newblock \bibinfo{journal}{Big Earth Data} \bibinfo{volume}{3},
  \bibinfo{pages}{232--254}.
\bibitem[{Spry and Teale(2021)}]{spry2021classification}
\bibinfo{author}{Spry, P.G.}, \bibinfo{author}{Teale, G.S.},
  \bibinfo{year}{2021}.
\newblock \bibinfo{title}{A classification of broken hill-type deposits: a
  critical review}.
\newblock \bibinfo{journal}{Ore Geology Reviews} \bibinfo{volume}{130},
  \bibinfo{pages}{103935}.
\bibitem[{Srivastava et~al.(2014)Srivastava, Hinton, Krizhevsky, Sutskever and
  Salakhutdinov}]{srivastava2014dropout}
\bibinfo{author}{Srivastava, N.}, \bibinfo{author}{Hinton, G.},
  \bibinfo{author}{Krizhevsky, A.}, \bibinfo{author}{Sutskever, I.},
  \bibinfo{author}{Salakhutdinov, R.}, \bibinfo{year}{2014}.
\newblock \bibinfo{title}{Dropout: a simple way to prevent neural networks from
  overfitting}.
\newblock \bibinfo{journal}{The Journal of Machine Learning Research}
  \bibinfo{volume}{15}, \bibinfo{pages}{1929--1958}.
\bibitem[{Stevens et~al.(1988)Stevens, Barnes, Brown, Stroud and
  Willis}]{stevens1988willyama}
\bibinfo{author}{Stevens, B.}, \bibinfo{author}{Barnes, R.},
  \bibinfo{author}{Brown, R.}, \bibinfo{author}{Stroud, W.},
  \bibinfo{author}{Willis, I.}, \bibinfo{year}{1988}.
\newblock \bibinfo{title}{The willyama supergroup in the broken hill and
  euriowie blocks, new south wales}.
\newblock \bibinfo{journal}{Precambrian Research} \bibinfo{volume}{40},
  \bibinfo{pages}{297--327}.
\bibitem[{Tahmasebi et~al.(2020)Tahmasebi, Kamrava, Bai and
  Sahimi}]{tahmasebi2020machine}
\bibinfo{author}{Tahmasebi, P.}, \bibinfo{author}{Kamrava, S.},
  \bibinfo{author}{Bai, T.}, \bibinfo{author}{Sahimi, M.},
  \bibinfo{year}{2020}.
\newblock \bibinfo{title}{Machine learning in geo-and environmental sciences:
  From small to large scale}.
\newblock \bibinfo{journal}{Advances in Water Resources} \bibinfo{volume}{142},
  \bibinfo{pages}{103619}.
\bibitem[{Taunk et~al.(2019)Taunk, De, Verma and Swetapadma}]{taunk2019brief}
\bibinfo{author}{Taunk, K.}, \bibinfo{author}{De, S.}, \bibinfo{author}{Verma,
  S.}, \bibinfo{author}{Swetapadma, A.}, \bibinfo{year}{2019}.
\newblock \bibinfo{title}{A brief review of nearest neighbor algorithm for
  learning and classification}, in: \bibinfo{booktitle}{International
  Conference on Intelligent Computing and Control Systems}, pp.
  \bibinfo{pages}{1255--1260}.
\newblock \DOIprefix\doi{10.1109/ICCS45141.2019.9065747}.
\bibitem[{Tawade and Virnodkar(2022)}]{tawade2022remote}
\bibinfo{author}{Tawade, A.}, \bibinfo{author}{Virnodkar, S.},
  \bibinfo{year}{2022}.
\newblock \bibinfo{title}{Remote sensing image fusion using machine learning
  and deep learning: a systematic review}, in: \bibinfo{booktitle}{7th
  International Conference on Computing in Engineering \& Technology}, pp.
  \bibinfo{pages}{36--46}.
\bibitem[{Vicente and de~Souza~Filho(2011)}]{vicente2011identification}
\bibinfo{author}{Vicente, L.E.}, \bibinfo{author}{de~Souza~Filho, C.R.},
  \bibinfo{year}{2011}.
\newblock \bibinfo{title}{Identification of mineral components in tropical
  soils using reflectance spectroscopy and advanced spaceborne thermal emission
  and reflection radiometer (aster) data}.
\newblock \bibinfo{journal}{Remote Sensing of Environment}
  \bibinfo{volume}{115}, \bibinfo{pages}{1824--1836}.
\bibitem[{Wang and Yeung(2020)}]{wang2020survey}
\bibinfo{author}{Wang, H.}, \bibinfo{author}{Yeung, D.Y.},
  \bibinfo{year}{2020}.
\newblock \bibinfo{title}{A survey on bayesian deep learning}.
\newblock \bibinfo{journal}{ACM Computing Surveys} \bibinfo{volume}{53},
  \bibinfo{pages}{1--37}.
\bibitem[{Wang et~al.(2018)Wang, Lu, Zhu, Lin and Wang}]{wang2018highspeed}
\bibinfo{author}{Wang, M.}, \bibinfo{author}{Lu, S.}, \bibinfo{author}{Zhu,
  D.}, \bibinfo{author}{Lin, J.}, \bibinfo{author}{Wang, Z.},
  \bibinfo{year}{2018}.
\newblock \bibinfo{title}{A high-speed and low-complexity architecture for
  softmax function in deep learning}, in: \bibinfo{booktitle}{IEEE Asia Pacific
  Conference on Circuits and Systems}, pp. \bibinfo{pages}{223--226}.
\newblock \DOIprefix\doi{10.1109/APCCAS.2018.8605654}.
\bibitem[{Wang et~al.(2020)Wang, Ma, Zhao and Tian}]{wang2020comprehensive}
\bibinfo{author}{Wang, Q.}, \bibinfo{author}{Ma, Y.}, \bibinfo{author}{Zhao,
  K.}, \bibinfo{author}{Tian, Y.}, \bibinfo{year}{2020}.
\newblock \bibinfo{title}{A comprehensive survey of loss functions in machine
  learning}.
\newblock \bibinfo{journal}{Annals of Data Science} , \bibinfo{pages}{1--26}.
\bibitem[{Williams et~al.(2005)Williams, Barton, Johnson, Fontboté, Haller,
  Mark, Oliver and Marschik}]{williams2005iron}
\bibinfo{author}{Williams, P.J.}, \bibinfo{author}{Barton, M.D.},
  \bibinfo{author}{Johnson, D.A.}, \bibinfo{author}{Fontboté, L.},
  \bibinfo{author}{Haller, A.d.}, \bibinfo{author}{Mark, G.},
  \bibinfo{author}{Oliver, N.H.}, \bibinfo{author}{Marschik, R.},
  \bibinfo{year}{2005}.
\newblock \bibinfo{title}{Iron oxide copper-gold deposits: Geology, space-time
  distribution, and possible modes of origin}, in: \bibinfo{booktitle}{One
  Hundredth Anniversary Volume}. \bibinfo{publisher}{Society of Economic
  Geologists}.
\bibitem[{Wu et~al.(2021)Wu, Hong and Chanussot}]{wu2021convolutional}
\bibinfo{author}{Wu, X.}, \bibinfo{author}{Hong, D.},
  \bibinfo{author}{Chanussot, J.}, \bibinfo{year}{2021}.
\newblock \bibinfo{title}{Convolutional neural networks for multimodal remote
  sensing data classification}.
\newblock \bibinfo{journal}{IEEE Transactions on Geoscience and Remote Sensing}
  \bibinfo{volume}{60}, \bibinfo{pages}{1--10}.
\bibitem[{Yang et~al.(2021)Yang, Hu and Stiefelhagen}]{yang2021context}
\bibinfo{author}{Yang, K.}, \bibinfo{author}{Hu, X.},
  \bibinfo{author}{Stiefelhagen, R.}, \bibinfo{year}{2021}.
\newblock \bibinfo{title}{Is context-aware cnn ready for the surroundings?
  panoramic semantic segmentation in the wild}.
\newblock \bibinfo{journal}{IEEE Transactions on Image Processing}
  \bibinfo{volume}{30}, \bibinfo{pages}{1866--1881}.
\bibitem[{Zhang and Zhang(2002)}]{zhang2002association}
\bibinfo{author}{Zhang, C.}, \bibinfo{author}{Zhang, S.}, \bibinfo{year}{2002}.
\newblock \bibinfo{title}{Association Rule Mining: Models and Algorithms}.
\newblock \bibinfo{publisher}{Springer}.
\bibitem[{Zhiqiang and Jun(2017)}]{zhiqiang2017review}
\bibinfo{author}{Zhiqiang, W.}, \bibinfo{author}{Jun, L.},
  \bibinfo{year}{2017}.
\newblock \bibinfo{title}{A review of object detection based on convolutional
  neural network}, in: \bibinfo{booktitle}{36th Chinese Control Conference},
  pp. \bibinfo{pages}{11104--11109}.
\bibitem[{Zhong et~al.(2016)Zhong, Gong and
  Sch{\"o}nlieb}]{zhong2016diversified}
\bibinfo{author}{Zhong, P.}, \bibinfo{author}{Gong, Z.},
  \bibinfo{author}{Sch{\"o}nlieb, C.}, \bibinfo{year}{2016}.
\newblock \bibinfo{title}{A diversified deep belief network for hyperspectral
  image classification}.
\newblock \bibinfo{journal}{International Archives of the Photogrammetry,
  Remote Sensing and Spatial Information Sciences} \bibinfo{volume}{41},
  \bibinfo{pages}{443--449}.
\bibitem[{Zweig and Campbell(1993)}]{zweig1993receiver}
\bibinfo{author}{Zweig, M.H.}, \bibinfo{author}{Campbell, G.},
  \bibinfo{year}{1993}.
\newblock \bibinfo{title}{Receiver-operating characteristic (roc) plots: a
  fundamental evaluation tool in clinical medicine}.
\newblock \bibinfo{journal}{Clinical Chemistry} \bibinfo{volume}{39},
  \bibinfo{pages}{561--577}.

\end{thebibliography}

\end{document}